\documentclass[10pt,twocolumn,letterpaper]{article}

\usepackage[pagenumbers]{cvpr} 

\usepackage{graphicx}
\usepackage{amsmath}
\usepackage{amssymb}
\usepackage{booktabs}
\usepackage{soul}
\usepackage{float}
\usepackage{gensymb}
\usepackage{cuted}
\usepackage{appendix}

\usepackage[pagebackref,breaklinks,colorlinks]{hyperref}

\usepackage[capitalize]{cleveref}
\crefname{section}{Sec.}{Secs.}
\Crefname{section}{Section}{Sections}
\Crefname{table}{Table}{Tables}
\crefname{table}{Tab.}{Tabs.}

\def\cvprPaperID{3022} 
\def\confName{CVPR}
\def\confYear{2022}

\title{VoLux-GAN: A Generative Model for 3D Face Synthesis with HDRI Relighting}

\author{Feitong Tan$^{1,2,}$ \thanks{Work done while the author was an intern at Google.} \ \ Sean Fanello$^{1}$ \ \ Abhimitra Meka$^{1}$ \ \ Sergio Orts-Escolano$^{1}$ \ \ Danhang Tang$^{1}$ \\ Rohit Pandey$^{1}$ \ \ Jonathan Taylor$^{1}$ \ \ Ping Tan$^{2}$ \ \ Yinda Zhang$^{1}$ \\
$^{1}$ Google  \ \ $^{2}$ Simon Fraser University
}

\begin{document}

\maketitle


\begin{strip}\centering
\vspace{-18mm}
\includegraphics[width=\textwidth]{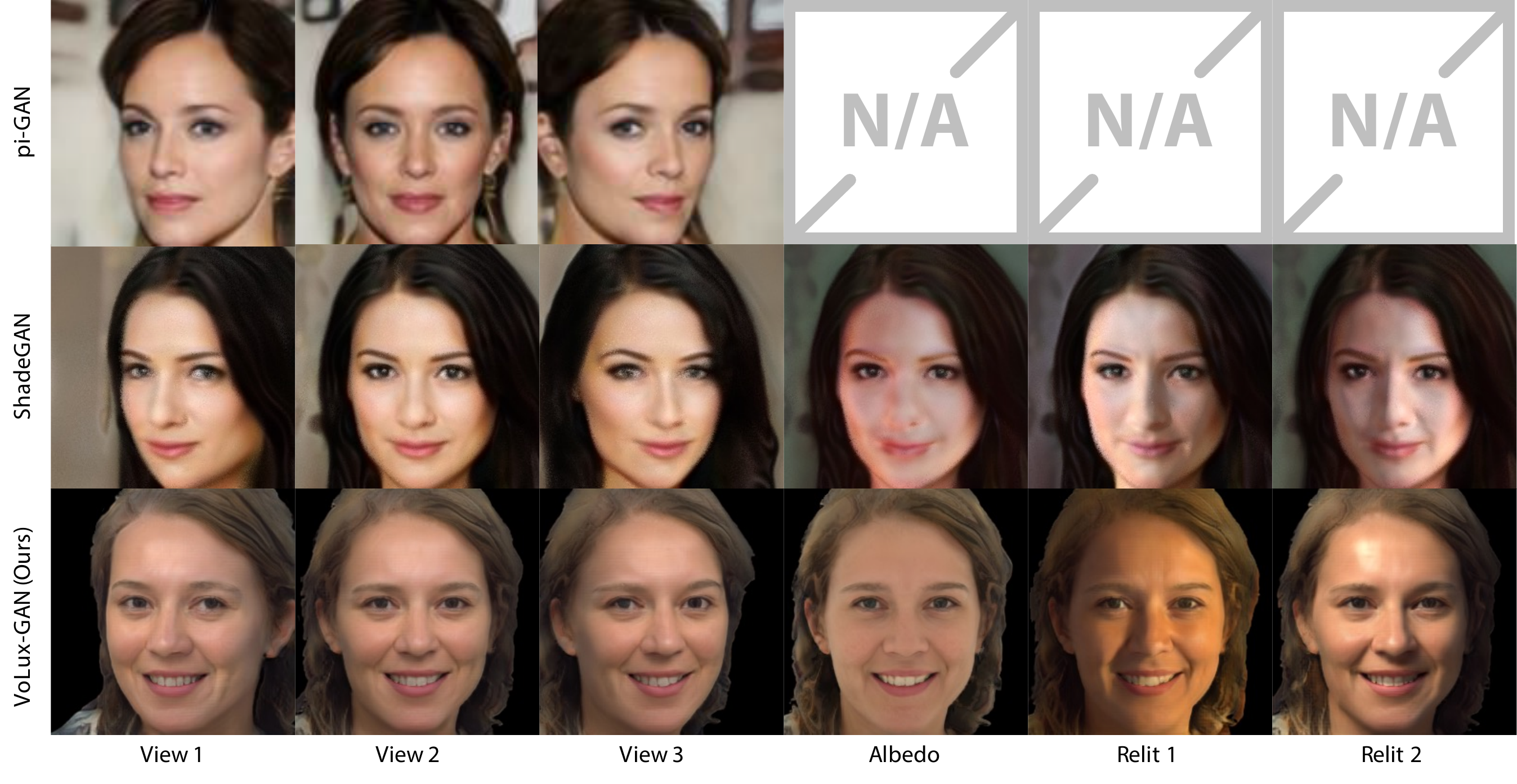}
\vspace{-8mm}
\captionof{figure}{We propose VoLux-GAN, a 3D-aware generator that produces faces with full HDRI relighting capability. Here we show a comparison of images generated by VoLux-GAN and related work pi-GAN \cite{chan2021pi} (which does not support relighting) and ShadeGAN \cite{pan2021shadegan}.}
\label{fig:teaser}
\end{strip}

\definecolor{MyDarkBlue}{rgb}{0,0.08,1}
\definecolor{MyCyan}{rgb}{0,0.6,0.6}
\definecolor{MyDarkGreen}{rgb}{0.02,0.6,0.02}
\definecolor{MyDarkRed}{rgb}{0.8,0.02,0.02}
\definecolor{MyDarkOrange}{rgb}{0.40,0.2,0.02}
\definecolor{MyPurple}{RGB}{111,0,255}
\definecolor{MyRed}{rgb}{1.0,0.0,0.0}
\definecolor{MyGold}{rgb}{0.75,0.6,0.12}
\definecolor{MyDarkgray}{rgb}{0.66, 0.66, 0.66}
\definecolor{MyOrange}{rgb}{1.0, 0.6, 0.0}
\definecolor{MyGreen}{rgb}{0.02,0.5,0.02}

\newcommand{\todelete}[1]{\textcolor{MyPurple}{[#1]}}
\newcommand{\sean}[1]{\textcolor{MyPurple}{[Sean: #1]}}
\newcommand{\todo}[1]{\textcolor{MyRed}{[TODO: #1]}}
\newcommand{\yinda}[1]{\textcolor{MyOrange}{[Yinda: #1]}}
\newcommand{\feitong}[1]{\textcolor{MyDarkBlue}{[Feitong: #1]}}
\newcommand{\jon}[1]{\textcolor{MyGreen}{[Jon: #1]}}
\newcommand{\rohit}[1]{\textcolor{MyCyan}{[Rohit: #1]}}
\newcommand{\abhi}[1]{\textcolor{MyDarkOrange}{[Abhi: #1]}}

\begin{abstract}
We propose VoLux-GAN, a generative framework to synthesize 3D-aware faces with convincing relighting. Our main contribution is a volumetric HDRI relighting method that can efficiently accumulate albedo, diffuse and specular lighting contributions along each 3D ray for any desired HDR environmental map. Additionally, we show the importance of supervising the image decomposition process using multiple discriminators. In particular, we propose a data augmentation technique that leverages recent advances in single image portrait relighting to enforce consistent geometry, albedo, diffuse and specular components. Multiple experiments and comparisons with other generative frameworks show how our model is a step forward towards photorealistic relightable 3D generative models.
\end{abstract}

\section{Introduction}
\label{sec:intro}
Generating synthetic novel human subjects with convincing photorealism is one of the most desired capabilities for automatic content generation and pseudo ground truth synthesis for machine learning. Such data generation engines can thus benefit many areas including the gaming and movie industries, telepresence in mixed reality, and computational photography. 
In order to achieve realism and flexibility when delivered in specific applications, the generated images should: 1) be enriched in details, \eg with high resolution; 2) support free viewpoint rendering to deliver immersive 3D experiences; 3) adapt to novel environmental illumination for realism; 4) synthesize novel identities for scalable data diversity. %

Motivated by these principles, in this paper we propose a neural human portrait generator, which delivers compelling rendering quality on arbitrary camera viewpoints and under any desired illumination.
With the success of Neural Radiance Field (NeRF) on volumetric rendering \cite{mildenhall2020nerf} and Generative Adversarial Networks (GAN) on image generation \cite{karras2019style}, 3D-aware generators \cite{gu2021stylenerf,chan2021pi,pan2021shadegan} have been proposed as a promising solution, which combine the merits of both.
By learning from a collection of portrait images, these methods are able to generate NeRF models from randomly sampled latent codes, which result in impressive free viewpoint rendering capabilities despite arguable underlying geometry quality and multi-view consistency.
Concurrent work proposed by Pan et al. \cite{pan2021shadegan} adds a shading model to enforce multi-lighting constraints during training, however the method shows substantial limitations in terms of photorealism and does not allow for full HDRI relighting.

In this work, we propose a 3D aware generative model with HDRI relighting supervised by adversarial losses. To overcome the limitations of prior arts, we identified and contributed to two main aspects:

\paragraph{Volumetric HDRI Relighting.}
We propose a novel approach of the volumetric rendering function that naturally supports efficient HDRI relighting. The core idea relies on the intuition that diffuse and specular components can be efficiently accumulated per-pixel when pre-filtered HDR lighting environments are used \cite{ramamoorthi2001efficient,greene1986environment}. This was successfully applied to single image portrait relighting \cite{pandey2021total}, and here we introduce an alternative formulation to allow for volumetric HDRI relighting.
Differently from \cite{pandey2021total,Wang20,nestmeyer2020learning} that predict surface normals and calculate the shading with respect to the light sources (for a given HDR environment map), we propose to directly integrate the diffuse and specular components at each 3D location along the ray according to their local surface normal and viewpoint direction. Simultaneously, an albedo image and neural features are accumulated along the 3D ray. Finally, a neural renderer combines the generated outputs to infer the final image.

\paragraph{Supervised Image Decomposition.}
Though producing impressive rendering quality, the geometry from 3D-aware generators is often incomplete or inaccurate \cite{chan2021pi,gu2021stylenerf}.
As a result, the model tends to bias the image quality for highly sampled camera views (e.g. front facing), but starts to show unsatisfactory multi-view consistency and 3D perception, breaking the photorealism when rendered from free-viewpoint camera trajectories. Additionally, high quality geometry is particularly important for relighting since any underlying reflectance models rely on accurate surface normal directions in order to correctly accumulate the light contributions from the HDR environment map.

Similarly, decomposing an image into albedo, diffuse and specular components without explicit supervision could lead to artifacts and inconsistencies, since, without any explicit constrains, the network could encode details in any channel even though it does not follow light transport principles.
For instance in Fig. \ref{fig:teaser}, the albedo image generated by previous methods \cite{pan2021shadegan} contains clear shading information, whereas the expected albedo (\ie flat lit image) should be closer to ours. At the same time, such supervision is  not available for in-the-wild datasets like FFHQ \cite{karras2019style}.

Motivated by this, and inspired by other works that apply pseudo-groundtruth labels \cite{chogovadze2021controllable} or synthetic renderings \cite{wood2021fake,tan2021humangps,saito2020pifuhd,zhu2020simpose} for in-the-wild tasks, we propose a data augmentation technique to explicitly supervise the image decomposition in geometry, albedo, diffuse and specular components. In particular, we employ the work of Pandey et al. \cite{pandey2021total} to generate albedo, geometry, diffuse, specular and relit images for each image of the dataset, and have additional discriminators guide the intrinsic decomposition during the training. This technique alone, however, would guide the generative model to synthesize images that are less photorealistic since their quality upper bound would depend on the specific image decomposition and relighting algorithm used as supervision (\eg \cite{pandey2021total}). In order to address this, we also add a final discriminator on the original images, which will guide the network towards real photorealism and higher order light transport effects such as specular highlights and subsurface scattering.

We summarize the contributions of this paper: 1) We propose a novel approach to generate HDRI relightable 3D faces with a volumetric rendering framework. 2) Supervised adversary losses are leveraged to increase the geometry and relighting quality, which also improves multi-view consistency. 3) Exhaustive experiments demonstrated the effectiveness of the framework for image synthesis and relighting. 


\section{Related Work}
\paragraph{2D Image Generation.}
Generating convincing renderings of humans is a very active trend in the field of neural rendering \cite{tewari2020state}. Here, we consider works that rely on a generative adversarial framework \cite{GANs} to synthesize photorealistic portraits. High quality results have been demonstrated by multiple early works \cite{durugkar2017generative,mordido2020dropoutgan,pmlr-v97-zhang19d} and since the groundbreaking work of StyleGAN \cite{karras2019style}, the community has made tremendous progress in synthesizing photorealistic and high resolution images  \cite{biggan,Karras2019stylegan2,Karras2021,choi2020stargan} with methods focusing on addressing most of the common issues with GANs including stability \cite{Karras2019stylegan2}, resolution \cite{biggan} and aliasing \cite{Karras2021}. These approaches generate impressive photorealistic images, but results typically lack free-viewpoint rendering and/or multi-view consistency.

\vspace{-4mm}
\paragraph{3D Aware Generation.}
Many recent approaches incorporated the use of geometry and its multi-view consistency to allow for 3D aware synthesis. \cite{geometric2018,chen2021ngp,visual2018,hologan19,phuoc2020blockgan,Liao2020CVPR,niemeyer2021giraffe,gu2021stylenerf,chan2021pi,zhou2021cips}. 
Past works rely on voxels \cite{hologan19,phuoc2020blockgan,visual2018,gadelha20173d}, meshes \cite{szabo2019unsupervised}, face models \cite{buehler2021varitex} or shape primitives \cite{Liao2020CVPR} as the 3D representation for image generation, but the majority have been limited to low resolution image generation.
Inspired by the success of NeRF \cite{mildenhall2020nerf}, methods \cite{schwarz2021graf,niemeyer2021giraffe,chan2021pi} adopt implicit volumetric rendering framework, and require only unconstrained images for 3D GAN training, but these architectures are computational consuming, which limit the training for high-resolution image generation. Concurrently, StyleNeRF \cite{gu2021stylenerf}, CIPS-3D\cite{zhou2021cips}, StyleSDF\cite{or2021stylesdf} adopt the two-stage rendering strategy to reduce the computation for high-resolution image generation. EG3D\cite{Chan2021} introduces tri-plane representation for fast and scalable rendering, and GRAM\cite{deng2021gram} proposes to render radiance manifolds first to produce high quality images. However, all these concurrent methods lack controllable relighting capabilities.

\vspace{-4mm}
\paragraph{Relightable Generative Models.}
Relightable NeRF models \cite{zhang2021nerfactor,Boss2020-NeRD,boss2021neuralpil,zhang2021ners} have shown that full image decomposition is possible when explicit multi-view imagery is provided as supervision. As for generative networks, the concurrent work of Pan et al. \cite{pan2021shadegan} is, to the best of our knowledge, the first at enabling relightability into generative model in a volumetrics 3D framework. 
The method enforces both multi-view and multi-lighting consistency to allow controllable viewpoint and illumination.
This approach, however, adopts a simplified Lambertian model and only supports one specific light direction at the time and extending it to full HDR relighting is computationally prohibitive. HeadNeRF\cite{hong2021headnerf} propose a NeRF-based parametric head model which can control the illumination by adjusting latent code. However, limited by insufficient coverage of illumination in their dataset, the method cannot control the image shading like continuously moving light source position. Also, they cannot explicitly control the illumination by adjusting latent code, while it is achievable in HDRI relighting by given a desired HDR map.

\vspace{-4mm}
\paragraph{Intrinsic Image Decomposition.} Decomposing an image into albedo, geometry and reflectance components has achieved using model-fitting techniques \cite{BarronM15,meka:2017} and deep learning based approaches \cite{Ren:2015,Meka:2018,xu2018deep,Kanamori:2018} that attempt at inferring image properties from one or multiple images. Very recently, state-of-art image based portrait relighting methods \cite{pandey2021total,Wang20,nestmeyer2020learning,tajima2021relighting} have shown impressive results by predicting explicit surface normals, albedo and shading information to formulate the interaction between light sources and geometry. These approaches usually rely on a specific shading model (e.g. Phong) and a neural renderer to synthesize the final image.

\vspace{-4mm}
\paragraph{Our Approach.} In contrast, we propose a volumetric generative model that supports full HDR relighting. We show how we can efficiently aggregate albedo, diffuse and specular components within the 3D volume. Thanks to the explicit supervision in our adversarial losses, we demonstrate that the method can perform such a full image component decomposition for novel face identities, starting from a randomly sampled latent code.

\begin{figure*}[t]
\centering
\includegraphics[width=0.95\linewidth]{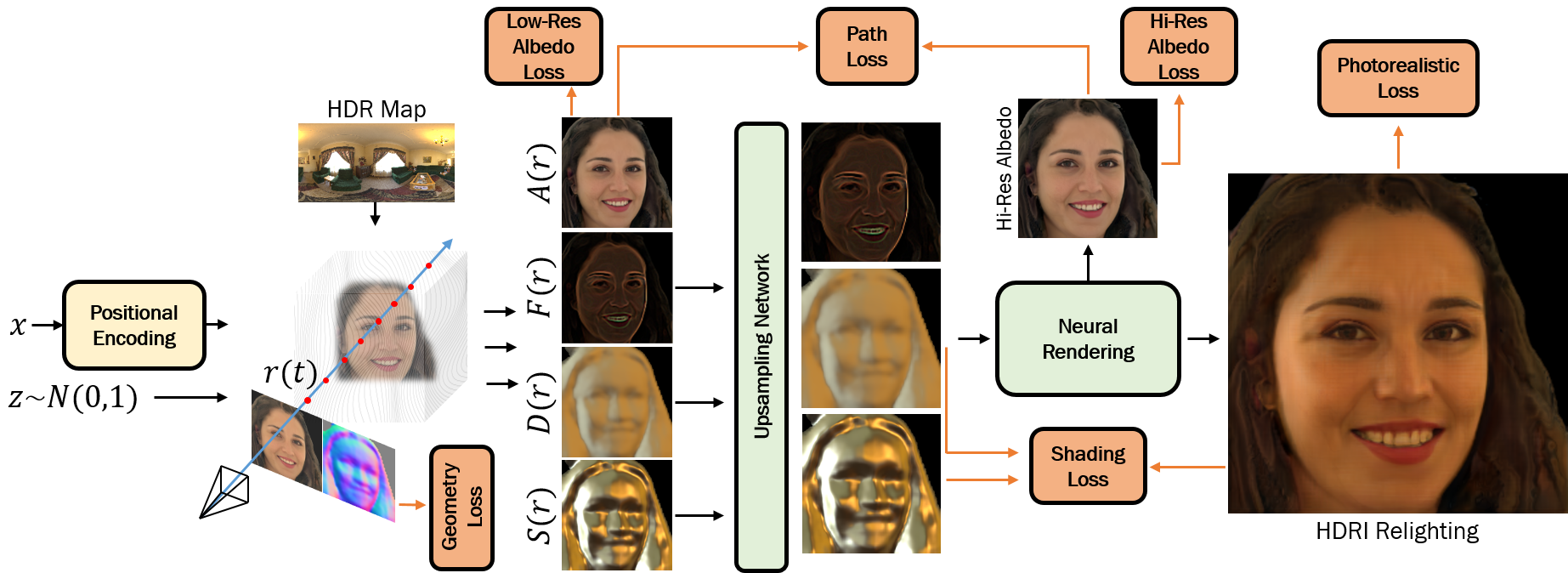}
\caption{VoLux-GAN Framework. Starting from a latent code we can efficiently accumulate albedo $A(\mathbf{r})$, surface normals $N(\mathbf{r})$, diffuse $D(\mathbf{r})$, specular components $S(\mathbf{r})$, and a feature map $F(\mathbf{r})$ along the 3D ray $\mathbf{r}(t)$ for any given HDR map. An upsampling strategy and a neural renderer synthesize the final relit image.}
\label{fig:framework}
\vspace{-15pt}
\end{figure*}

\section{VoLux-GAN Framework}
In this section, we introduce our neural generator that produces novel faces that can be rendered at free camera viewpoints and relit under an arbitrary HDR environment light map.
Our method starts from a neural implicit field that takes a randomly sampled latent vector as input and produces an albedo, volume density, and reflectance properties for any queried 3D location.  These outputs are then aggregated via volumetric rendering to produce low resolution albedo, diffuse shading, specular shading, and neural feature maps.
These intermediate outputs are then upsampled to high resolution and fed into a neural renderer to produce relit images. The overall framework is depicted in Figure \ref{fig:framework}.

\subsection{Preliminaries: Neural Volumetric Rendering.}
To aid the reader, we first briefly introduce the neural volumetric rendering framework originally presented in Mildenhall \etal \cite{mildenhall2020nerf}.
There, the 3D appearance of an object of interest is encoded into a neural implicit field implemented using a multilayer perceptron (MLP), which takes a 3D coordinate $x \in R^3$ (mapped through a sinusoidal function based positional encoding \cite{mildenhall2020nerf, tancik2021})
and viewing direction $\textbf{d} \in S^2$ as inputs and outputs a volume density $\sigma \in R^+$ and view-dependent color $\textbf{c} \in R^3$.
To render an image, the pixel color $\textbf{C}$ is accumulated along each camera ray $\textbf{r}(t)=\textbf{o}+t\textbf{d}$ as
\begin{equation}
    \mathbf{C}(\mathbf{r},\mathbf{d})=\int_{t_n}^{t_f} T(t) \sigma(\mathbf{r}(t)) \mathbf{c}(\mathbf{r}(t),\mathbf{d}) dt,
    \label{eq:nerf}
\end{equation}
where $T(t)=exp(-\int_{t_n}^{t}  \sigma(\textbf{r}(s))ds)$ and bounds $t_n$ and $t_f$.
Compared to surface based rendering, volumetric rendering more naturally handles translucent materials and regions with complex geometry such as thin structures.

\subsection{Generative Neural Implicit Intrinsic Field.}
Similar to other state-of-the-art 3D-aware generators \cite{chan2021pi, gu2021stylenerf, pan2021shadegan}, we train a MLP-based neural implicit field conditioned on a latent code $z$ sampled from a Gaussian distribution $N(0,I)^d$ and extend it to support HDRI relighting.
We adopt a Phong shading model \cite{phong1975illumination}, where the illumination of each point is determined by albedo, diffuse, and specular component.
Therefore, instead of having the network predict per-point radiance and directly obtaining a color image (via Eq. \ref{eq:nerf}), our network produces per-point albedo ($\alpha$), density ($\sigma$) and reflectance properties from separate MLP heads. 
The normal directions are obtained via the spatial derivative of the density field, which are used together with HDR illumination to compute diffuse and specular shading. Similar to \cite{pandey2021total}, rather than explicitly using the Phong model for the final rendering, we feed the albedo, diffuse and specular components to a lightweight neural renderer, which can also model higher order light transport effects.

\vspace{2mm}
\noindent\textbf{Efficient Shading Computation.}
Concurrent work \cite{pan2021shadegan} assumes Lambertian shading from a single light source.
Extending this to support full HDR illumination would require the integration of the shading contribution from multiple positional lights, making the approach computationally prohibitive, especially when performed at training time for millions of images. 
Inspired by the success of recent image based portrait relighting work \cite{pandey2021total}, we adopt a method designed for real-time shading rendering under HDR illumination \cite{greene1986environment,ramamoorthi2001efficient}.
The core idea is to approximate the diffuse and specular components using a preconvolved HDRI map.
Specifically, we first preconvolve the given HDRI map ($\mathbf{H}$) into light maps ($L_{n_i}, i=1,2,\cdots,N$) with cosine lobe functions corresponding to a set of pre-selected Phong specular exponents ($n_i, i=1,2,\cdots,N$) \cite{miller1984illumination}.
The diffuse shading $D$ is the first light map (\ie $n=1$ above) following the surface normal direction, and the specular shading is defined as a linear combination of all light maps indexed by the reflection direction. In order to capture possible diverse material properties of the face, we let the network estimate the blending weights ($\omega_i$) with another MLP branch, which are then used for the specular component $S$.

\vspace{2mm}
\noindent\textbf{Volumetric Shading Rendering.}
Typically, a reflectance model is defined on a surface \cite{phong1975illumination} and relighting methods \cite{pandey2021total,nestmeyer2020learning,Wang20} explicitly estimate surface normals from a single image.
Here, we propose a volumetric formulation to compute albedo, diffuse and view-dependent specular shading maps as:
\begin{align}
\label{eq:volux}
\begin{split}
    A(\mathbf{r}) &=\int_{t_{n}}^{t_{f}}T(t)\sigma\left(\mathbf{r}(t)\right)\alpha(\mathbf{r}(t)) dt 
    \\
    D(\mathbf{r}) &=\int_{t_{n}}^{t_{f}}T(t)\sigma\left(\mathbf{r}(t)\right)L_{n=1}(\mathbf{n}\left(t\right)) dt 
     \\
    S(\mathbf{r}) &=\int_{t_{n}}^{t_{f}}T(t)\sigma\left(\mathbf{r}(t)\right)\sum_{i}^{N}\omega_{i}L_{n_{i}}(\mathbf{n}(t), \mathbf{d}) dt
    \\
    F(\mathbf{r}) &=\int_{t_{n}}^{t_{f}}T(t)\sigma\left(\mathbf{r}(t)\right)f(\mathbf{r}(t)) dt 
\end{split}
\end{align}
where $\mathbf{n}(t)$ is the normal direction estimated via $\nabla \sigma(\mathbf{r}(t))$, $L_{n=1}(\mathbf{n}\left(t\right))$ is the diffuse light map indexed by the normal direction $\mathbf{n}(t)$, and $L_{n_{i}}(\mathbf{n}(t), \mathbf{d})$ is the specular component $n_i$ indexed by the inbound reflection direction depending on the local normal and viewing direction $\mathbf{d}$. 
Finally, $\alpha,\sigma,\omega$, and a per-location feature $f$ are the network outputs conditioned on the sampled latent code $z$.
We restrict the albedo to be view and lighting independent and encourage multi-view consistency. Note that in addition to rendering components such as albedo, diffuse and specular components, we let our network accumulate additional features $F(\mathbf{r})$, so that it can capture high frequency details and material properties in an unsupervised fashion.

\vspace{2mm}
\noindent\textbf{Volumetric Model Network Architecture}
We extend the architecture from concurrent work proposed by Gu \etal \cite{gu2021stylenerf} for our neural implicit field.
Rather than explicitly use the low resolution albedo $A(\mathbf{r})$ following Eq. \ref{eq:volux}, our network produces a feature vector ${f(\mathbf{r}(t))} \in R^{256}$ via 6 fully-connected layers from the positional encoding on the 3D coordinates.
A linear-layer is attached to the output of the 4-th layer to produce the volume density, and an additional two-layer MLP is attached to 6-th layer to produces the albedo and reflectance properties. Diffuse component $D$ and Specular Component $S$ are estimated following Eq. \ref{eq:volux}, where the blending weights $\omega_i$ are estimated by the network.  

\vspace{2mm}
\noindent\textbf{Neural Rendering Network}
\label{sec:relit_network}
To reduce the memory consumption and computation cost, we render albedo, diffuse, and specular shading in low resolution and upsample them to high resolution for relighting. The specific low and high resolutions depend on the dataset used and details can be found in the Section \ref{sec:experiments}.
To generate the high resolution albedo, we upsample the feature map $F(\mathbf{r})$ and enforce it’s
first 3 channels to correspond to the albedo image, similar to some other works in the literature \cite{deep_relightables,deferred}.
Each upsampling unit consists of two $1 \times 1$ convolutions modulated by the latent code $z$, a pixelshuffle upsampler \cite{shi2016real}  and a BlurPool \cite{zhang2019making} with stride $1$. The low resolution albedo $A(\mathbf{r})$ is still used to enforce consistency with the upsampled high resolution albedo (see Section \ref{sec:loss}).
For shading maps, we directly apply bilinear upsampling.


Finally, a relighting network takes as input the albedo map $A$, the diffuse map $D$, the specular component map $S$ and the features $F$ and generates the final $I_{relit}$ image.
The architecture of Relighting Network is a shallow U-Net \cite{ronneberger2015u}.

\subsection{Supervised Adversarial Training}
\label{sec:loss}
Here we introduce the scheme to train our pipeline from a collection of unconstrained in-the-wild images.
While it is possible to train the full pipeline with a single adversarial loss on the relit image, we found empirically that adding additional supervision on intermediate outputs significantly improves the training convergence and rendering quality.

\begin{figure}
\centering
\vspace{-2mm}
\includegraphics[width=\linewidth]{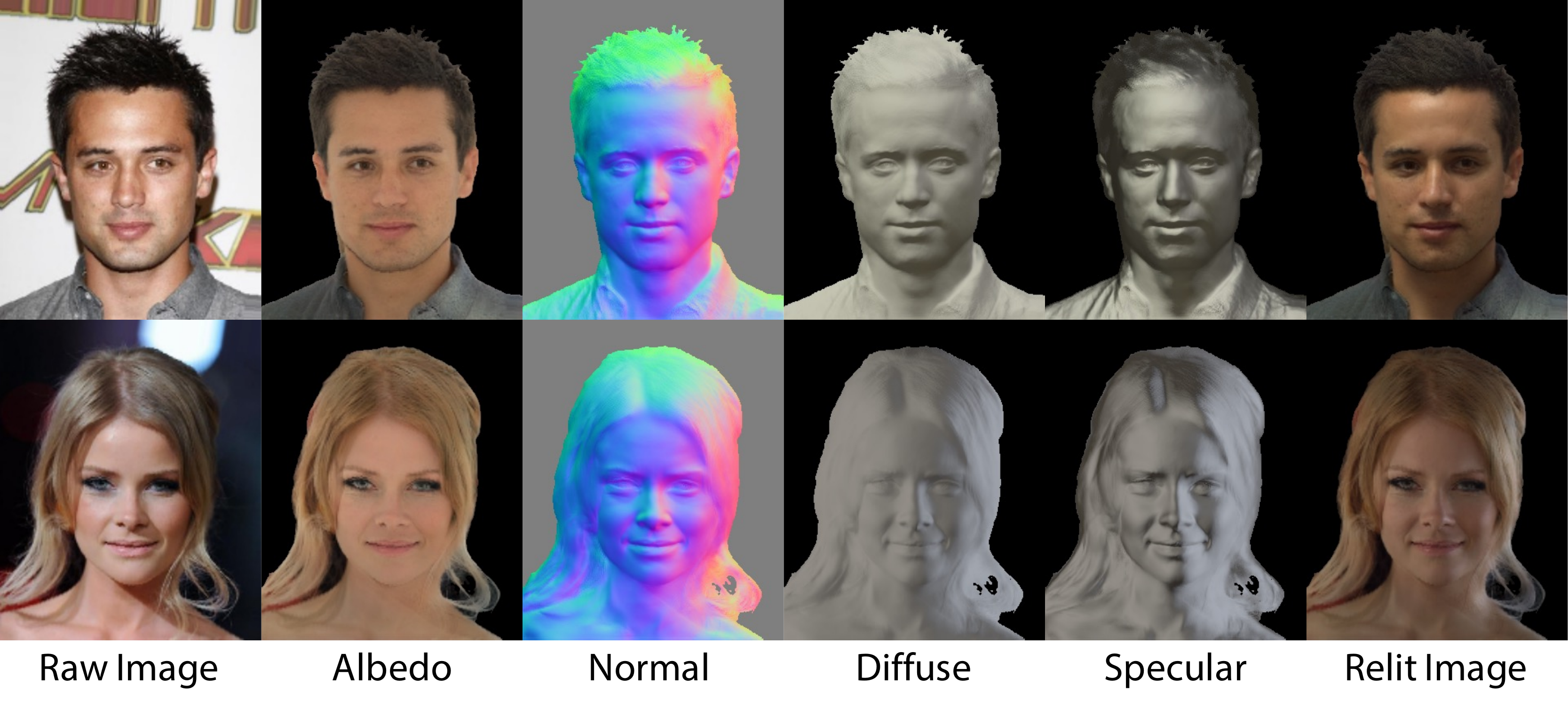}
\vspace{-8mm}
\caption{Relighting augmentation on CelebA \cite{liu2015faceattributes}. We run Total Relighting \cite{pandey2021total} to generate albedo, normal, shading, and relit images, which supervise the training via adversarial losses.}
\label{fig:dataset}
\vspace{-6mm}
\end{figure}

\vspace{2mm}
\noindent\textbf{Pesudo Ground Truth Generation.}
Large scale in-the-wild images provides great data diversity, which is critical for training a generator.
However, the groundtruth labels for geometry and shading are usually missing.
In our case, we are particularly interested in having ``real examples'' of the albedo and geometry to supervise our method.
To this end, we resort to the state-of-the-art image based relighting algorithm, Total Relighting \cite{pandey2021total}, to produce pseudo ground truth albedo and normals and to also further increase the data diversity. 
Specifically, for each image in our training set, we randomly select an HDRI map from a collection of $400$ maps sourced from public repository \cite{HDRIHaven}, apply a random rotation, and run Total Relighting to generate the albedo, surface normal and a relit image with the associated light maps (diffuse and specular components). Example images from the CelebA dataset \cite{liu2015faceattributes} augmented with this technique are shown in Figure \ref{fig:dataset}.

\vspace{2mm}
\noindent\textbf{Albedo Adversarial Loss $\mathcal{L}_{A}$: $D_{A}(A(\mathbf{r}))+D_{A}(A_{hi-res})$} 
We supervise the output albedo images in both low and high resolution with adversarial loss using the pseudo ground truth generated with \cite{pandey2021total}.
A standard non-saturating logistic GAN loss with R1 penalty is applied to train the generator and discriminator. The discriminator architecture $D_*$ for all the losses follows the one proposed in \cite{Karras2019stylegan2}.

\vspace{2mm}
\noindent\textbf{Geometry Adversarial Loss $\mathcal{L}_{G}$: $D_{G}(\nabla \sigma(\mathbf{r}(t)))$}
We also supervise the geometry as it is crucial for multi-view consistent rendering and relighting realism.
While the density $\sigma$ is the immediate output from the network that measures the geometry, we find it is more convenient to supervise the surface normals computed via $\nabla \sigma(\mathbf{r}(t))$. Therefore, we add an adversarial loss between the volumetric rendered normal from the derivative of the density and the pseudo ground truth normal obtained from \cite{pandey2021total}.

\vspace{2mm}
\noindent\textbf{Shading Adversarial Loss $\mathcal{L}_{S}$: $D_{S}(D(\mathbf{r}), S(\mathbf{r}), I_{relit})$} To enforce that the Relight Network faithfully integrates shading with albedo, we apply a conditional adversarial loss on the relit image.
This is achieved by adding a discriminator $D_S$ that takes the concatenation of the relit image $I_{relit}$, diffuse map $D(\mathbf{r})$ and specular map $S(\mathbf{r})$ as the inputs and discriminate if the group is fake, \ie from our model, or true, \ie from \cite{pandey2021total}.
The training gradients are only allowed to back-propagate to the relit image but not the other inputs (\ie set to zero) as they are the data to be conditioned on. 

\vspace{2mm}
\noindent\textbf{Photorealistic Adversarial Loss $\mathcal{L}_{P}$: $D_{P}(I_{relit})$}
A downside of the Shading Adversarial Loss is that the model performance is upper-bounded by the specific algorithm used to generate pseudo-groundtruth labels, in our case \cite{pandey2021total}.
As a result, inaccuracies in the relighting examples, \eg overly smoothed shading and lack of specular highlights, may affect our rendering quality.
To enhance the photorealism, we add one final adversarial loss directly on the generated relit images with the original images from the dataset.

\vspace{2mm}
\noindent\textbf{Path Loss $\mathcal{L}_{path}$: $\ell_{1}(A(\mathbf{r}),A_{hi-res})$}
Following StyleNeRF \cite{gu2021stylenerf}, we add a path loss to ensure the consistency between the albedo maps in low and high resolutions.
Specifically, we downsample the high resolution to the low resolution, and add a per-pixel $\ell_1$ loss.

The final loss function is a weighted sum of all above mentioned terms: $\mathcal{L}=\lambda_1 \mathcal{L}_{A}+ \lambda_2 \mathcal{L}_{G}+ \lambda_3 \mathcal{L}_{S}+ \lambda_4 \mathcal{L}_{P}+ \lambda_5 \mathcal{L}_{path}$, where for our experiments we empirically determined these weights to be $1.0, 0.5, 0.25, 0.75, 0.5$.

\begin{figure*}
\centering
\vspace{-3mm}
\includegraphics[width=\textwidth]{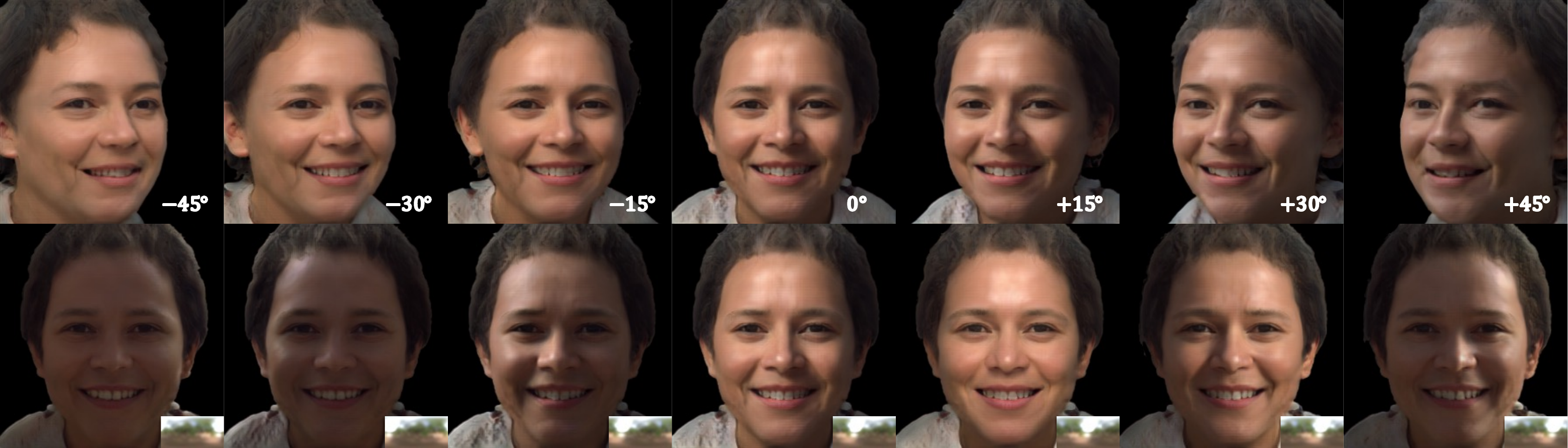}
\vspace{-15pt}
\caption{Our synthesized images under rotating camera or rotating lighting. Note the relighting consistency and view-dependent effects.}
\vspace{-1mm}
\label{fig:rotating}
\end{figure*}

\begin{figure*}
\centering
\includegraphics[width=\textwidth]{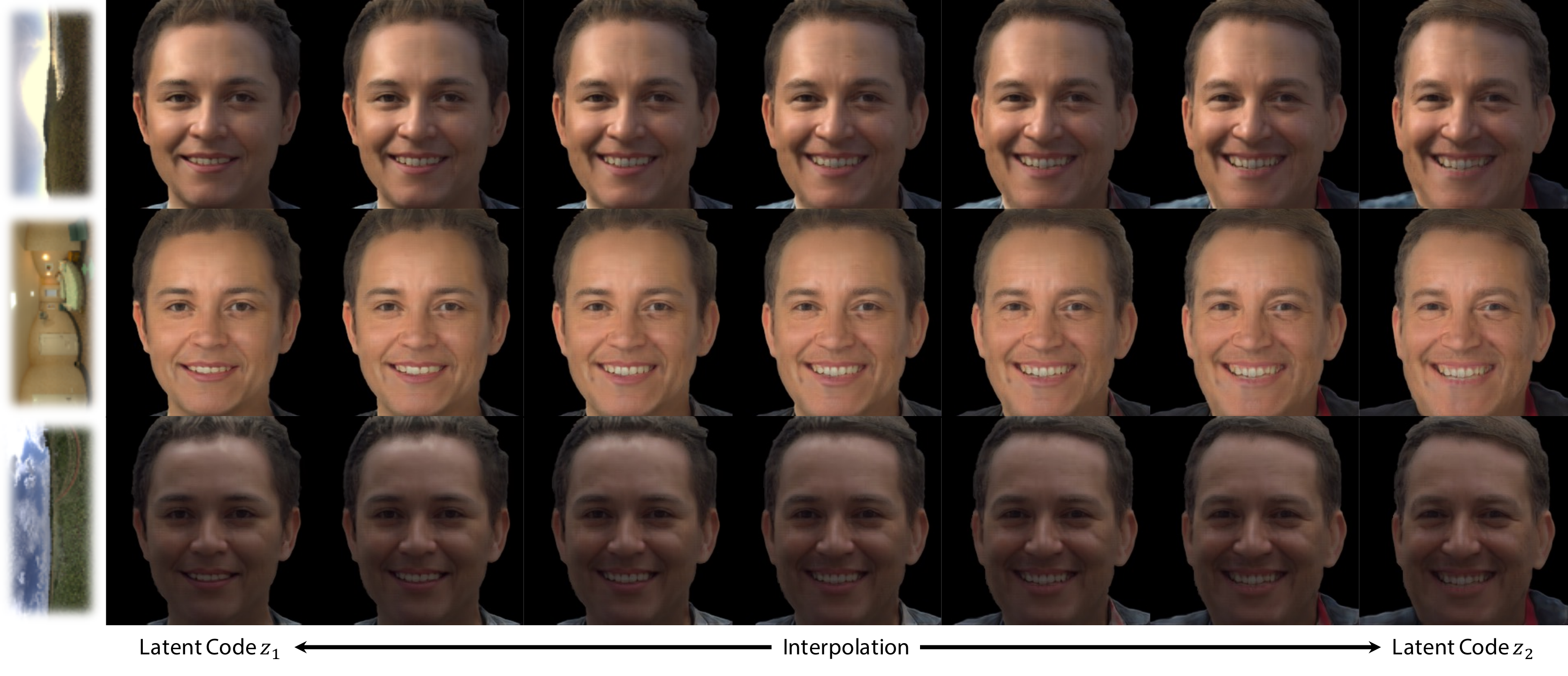}
\vspace{-20pt}
\caption{Our result with under interpolated latent code and relighting results. Note the smooth transition across users where relighting effects are correctly transferred.}
\label{fig:interpolation}
\vspace{-10pt}
\end{figure*}

\subsection{Implementation Details}
To ensure stable training, we first train the neural implicit field and the upsampling network for high quality albedo and geometry.
We only enable albedo and geometry adversarial loss and adopt progressive growing training strategy \cite{karras2017progressive} with the path loss to train the upsampling network. 
Once the network converges, we enable all the loss terms and train the whole network end-to-end.
For the optimization, we use the Adam optimizer with $\beta_1=0$, $\beta_2=0.99$. The batch size is set as $24$ and the learning rates for the generator and the discriminator are set empirically to $0.0022$ and $0.0025$ respectively. We train the VoLux-GAN model for $1$ million iterations with albedo and geometry adversarial losses, then it is trained for additional $500$k iterations to generate the relit images.

\section{Experiments}
\label{sec:experiments}
In this section, we compare our rendering quality and relighting performance with state-of-the-art methods.
We also provide ablation study showing the contribution of major system design choice to the final performance.

\vspace{1mm}
\noindent\textbf{Datasets.}
We train our model on CelebA dataset \cite{liu2015faceattributes} which is widely used for such comparisons, and on the FFHQ \cite{karras2019style} where a comparison of high resolution results can be made.
On CelebA, our model produces volumetric renderings at $64\times64$ and final outputs at $128\times128$.
On FFHQ, the volumetric renderings and final resolution are $64\times64$ and $256\times256$ respectively.

\vspace{1mm}
\noindent\textbf{Baseline Methods.}
We show qualitative and quantitative comparison with ShadeGAN \cite{pan2021shadegan} since, to the best of our knowledge, it is the only 3D-aware generator that supports relighting.
In addition, we consider an alternative strong baseline where we use pi-GAN \cite{chan2021pi} to render multi-view images and then run a single image based portrait relighting method \cite{pandey2021total} for HDR relighting.

\vspace{1mm}
\noindent\textbf{Metrics.}
Many evaluation metrics relying on perception features have been proposed to measure the rendering quality \cite{heusel2017gans, binkowski2018demystifying}.
While these metrics indeed measure the similarity between two collections of images, they are very sensitive to implementation details such as training image resolution, image post-processing \eg cropping, or the choice of training dataset, as has been shown in literature \cite{parmar2021cleanfid}.
As a result, these metrics are not suitable to evaluate our model since our pipeline is not trained directly on publicly available datasets but on a specifically tailored augmented dataset for good rendering and relighting performance.

Instead, we evaluate our pipeline and other methods by measuring the perceptual impact of view-synthesis and relighting on a subject's identity.
Specifically, we use a similarity metric based on the embedding space of a state-of-the-art face recognition network\cite{deng2019arcface}. This stability metric indicates how well the subject identity is preserved when we synthesize novel views and novel light renderings for a synthesized face.

\subsection{Relightable Face Generation}
In this section we demonstrate the capabilities of our framework. In Fig. \ref{fig:rotating}, we show one subject randomly sampled from latent space $z~N(0, I)$ trained on the FFHQ dataset.
The first row shows the faces rendered at different camera poses.
Our network successfully renders consistent faces even under a large yaw angle (\eg $45\degree$) thanks to better geometry supervised by the geometry adversarial loss.
The second row shows the same subject rendered under a rotating HDR map.
Note how the specularities and shading on the face respond correctly to the HDR environment maps.

Our latent space also supports interpolation.
In Fig. \ref{fig:interpolation}, we linearly interpolate between two randomly sampled latent codes, and show relit images of each subject under three HDR lighting conditions.
As seen, the appearance of the subject transitions smoothly and the intermediate identities are successfully relit. Note also the consistent relighting, where view dependent effects and specularities are successfully transferred between different latent codes.

\subsection{Comparisons with State-of-the-Art}
We compare to ShadeGAN \cite{pan2021shadegan}, pi-GAN \cite{chan2021pi} coupled to an image based relighting \cite{pandey2021total}, and show the qualitative results in Fig. \ref{fig:compare_shadegan}. For a fair comparison in terms of image resolution, we trained our model on CelebA, like ShadeGAN and pi-GAN.
In each row, we show the albedo map and color images rendered under two different lightings from three camera viewpoints.

Our method produces significantly better albedo than ShadeGAN thanks to our supervised albedo adversarial loss.
Moreover, our results contain more high frequency specular components and can respond to more diverse global illumination.
pi-GAN coupled with \cite{pandey2021total} can produce plausible relighting results but occasionally with inconsistent shading (\eg in the 3rd row middle, the cheek is dark in one view but bright in another).
In contrast, our results show more consistency across views and lights thanks to the proposed volumetric relighting formulation, which is also reflected by the quantitative metric below.

\subsubsection{Quantitative Results}
We evaluate our method with quantitative metrics. The goal of the following experiments is to demonstrate that our method is able to synthesize images that are consistent across views in terms of geometry and relighting.

\vspace{2mm}
\noindent\textbf{Geometry Consistency.}
To demonstrate the geometry consistency, we render a fixed random latent code to multiple views. We then compute the similarity score \cite{deng2019arcface} of yaw-posed renderings with the frontal facing rendering, and average it over $100$ such randomly sampled latent codes. We compute the score for both relit images and the high-res albedo images. We also show the score computed with the same scheme for ShadeGAN \cite{pan2021shadegan} and the baseline of pi-GAN \cite{chan2021pi} + portrait relighting \cite{pandey2021total}.
The results are showed in Table \ref{tbl:view_consistency}.\
Note how our method consistently outperforms the other state-of-art approaches, demonstrating better multi-view consistency for each identity on generated albedo and relit images.

\vspace{2mm}
\noindent\textbf{Relighting Consistency.}
Similarly, we also evaluate the stability of our relighting. Following the geometry consistency experiment, we use the embedding space of a face recognition network~\cite{deng2019arcface} to generate the identity similarity score between the albedo and $3$  renderings under different environment maps (as shown in Fig. \ref{fig:interpolation}). We report an average score over $100$ randomly sampled latent codes. A stable relighting method should give a high similarity score, since relighting does not change the identity.

\begin{figure*}
\centering
\vspace{-2mm}
\includegraphics[width=0.98\textwidth]{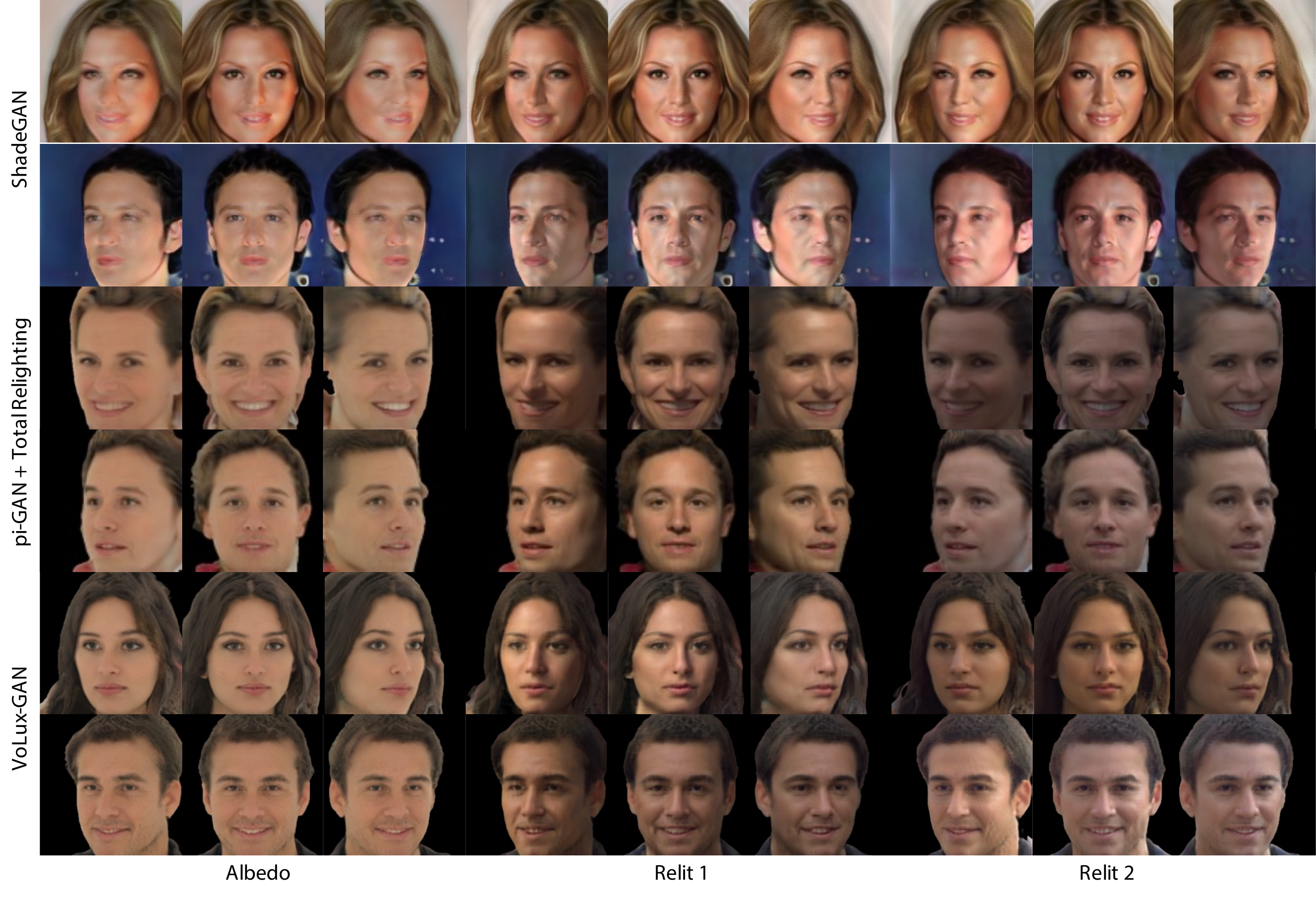}
\vspace{-15pt}
\caption{Qualitative comparisons on CelebA with \cite{pan2021shadegan} and pi-GAN \cite{chan2021pi} + portrait relighting \cite{pandey2021total}: note how our method produces more consistent albedo and relighting results across multiple views.}
\label{fig:compare_shadegan}
\end{figure*}

The results are reported in Table \ref{tab:relighting_stability}, showing that our approach is also able to generate relit images that are more consistent with the original albedo identity. At the same time our relit images look more photorealistic as shown in Figure \ref{fig:compare_shadegan}, where we better capture higher order light transport effects.

\begin{table*}
\small
\begin{center}
\resizebox{\textwidth}{!}{%
\begin{tabular}{|c | ccccc|ccccc |}
\hline
Method & \multicolumn{5}{c}{\textbf{Relit image identity similarity$\uparrow$}} & \multicolumn{5}{c|}{\textbf{Albedo image identity similarity$\uparrow$}} \\
\cline{2-11}
& -0.5 rad & -0.25 rad & 0 rad & 0.25 rad & 0.5 rad & -0.5 rad & -0.25 rad & 0 rad & 0.25 rad & 0.5 rad \\
\hline

ShadeGAN\cite{pan2021shadegan} & 0.4814 & 0.7513 & - & 0.7628 &  0.4997 & 0.4818 & 0.7582 & - &  0.7702 & 0.5091 \\
pi-GAN \cite{chan2021pi} + TR\cite{pandey2021total} & 0.5215 & 0.7472 & - & 0.7419 & 0.4981 &   0.5135 & 0.7378 & -& 0.7438 & 0.4898 \\

\hline
VoLux-GAN+Surface Relighting & 0.4471 & 0.7065 & - & 0.7388 & 0.4959 & 0.5531 & 0.7611 & - & 0.7796 & 0.5585 \\
VoLux-GAN - $\mathcal{L}_{P}$ & 0.4827 & 0.6487 & - & 0.6721 & 0.5156 & 0.5467 & 0.7809 & - & 0.8151 & 0.5901 \\
VoLux-GAN - $\mathcal{L}_{S}$ & 0.4071 & 0.6996 & - & 0.7788 & 0.4718 & 0.4886 & 0.7292 & - & 0.8095 & 0.5581 \\
VoLux-GAN - $\mathcal{L}_{G}$ & 0.4776 & 0.7182 & - & 0.7564 & 0.5015 & 0.4652 & 0.7278 & - & \textbf{0.8099} & 0.5398 \\
\hline
\textbf{VoLux-GAN} & \textbf{0.6064} & \textbf{0.7736} & - & \textbf{0.7997} & \textbf{0.5985} & \textbf{0.6389} & \textbf{0.7919} & - & 0.7863 & \textbf{0.6162} \\
\hline
\end{tabular}
}
\vspace{-15pt}
\end{center}
\caption{Identity consistency across camera poses around the yaw axis. The scores indicate the similarity calculated as the dot product between normalized embeddings from a state-of-the-art face recognition network~\cite{deng2019arcface} (higher is better). While our method performs comparably or marginally better at small view changes, we significantly outperform the state-of-the-art at more extreme viewpoints. \label{tbl:view_consistency}}
\vspace{-10pt}
\end{table*}

\begin{table}
\centering
\resizebox{\columnwidth}{!}{%
\begin{tabular}{|c|cccc|}
\hline
Method & HDR map 1 & HDR map 2 & HDR map 3 & \tabularnewline
\hline
ShadeGAN\cite{pan2021shadegan} &  0.5806 & 0.6486 & 0.6559 & \tabularnewline
pi-GAN\cite{chan2021pi} + TR\cite{pandey2021total} & 0.7014 & 0.8796  & 0.7677  & \tabularnewline
\hline
VoLux-GAN+Surface Relighting &  0.5510 &  0.5548 & 0.5349 & \tabularnewline
VoLux-GAN - $\mathcal{L}_{P}$ &  0.4132 &  0.4471 & 0.4712 & \tabularnewline
VoLux-GAN - $\mathcal{L}_{S}$ &  \textbf{0.8917} &   0.8846 & 0.7387 & \tabularnewline
VoLux-GAN - $\mathcal{L}_{G}$ &  0.5331 &  0.5864 & 0.6158 & \tabularnewline
\hline
VoLux-GAN &  0.7600 &  \textbf{0.8900} & \textbf{0.8082} & \tabularnewline
\hline
\end{tabular}%
}
\vspace{-5pt}
\caption{Relighting consistency. Our method is able to better preserve the original identity (albedo) under different illuminations. Note that ShadeGAN is evaluated over three different positional light sources instead of HDR maps.}
\label{tab:relighting_stability}
\vspace{-15pt}
\end{table}

\vspace{2mm}
\noindent\textbf{Ablation Study.}
We report an ablation study of the loss functions used for training out pipeline and Table \ref{tab:relighting_stability}. As demonstrated by these quantitative results, the full framework and the proposed supervised adversarial losses are all contributing to the final rendering quality. In particular, we show that removing the shading loss $\mathcal{L}_{S}$, or geometry loss $\mathcal{L}_{G}$ or photorealistic $\mathcal{L}_{P}$ all lead to lower metrics. Similarly, when we first accumulate the surface geometry and then perform image based relighting, the results are unsatisfactory (see comparison with VoLux-GAN+Surface Relighting in Table \ref{tbl:view_consistency} and Table \ref{tab:relighting_stability}).


%

\section{Discussion}
We proposed a generative model of face images, that internally leverages a volumetric representation to facilitate multi-view generation and full HDRI relighting. 
Of particular note is that we have shown how to efficiently perform the aggregation of albedo, specular and diffuse components that helps to preserve the identity. 
Furthermore, a proposed supervised adversarial framework guides the network to generate the right intrinsic properties of faces. Our results prove the effectiveness of the approach for synthesizing novel identities. Future work could explore the use of semantic information to allow for expression control similar to StyleGAN \cite{karras2019style}.
While far from the intent of this work, we do recognize that generative models could be misused to synthesize fake content (see \cite{tolosana20} for an exhaustive survey). We believe that in order to address this, the community should prioritize open-sourcing pre-trained models to encourage the development of forgery detection methods. Great steps in that direction have been made thanks to the availability of datasets such as FaceForensics++ \cite{roessler2019faceforensics}. Additionally, in order to mitigate misuses, researchers could put more emphasis on the adversarial models (\ie discriminators) and make them publicly available when releasing a generative model. Enforcing higher importance to the discriminator loss, while fixing the generator could provide an effective method to detect misuse of the specific generative model.

{\small
\bibliographystyle{ieee_fullname}
\bibliography{egbib}

\begin{thebibliography}{10}\itemsep=-1pt

\bibitem{10.1145/3447648}
Rameen Abdal, Peihao Zhu, Niloy~J. Mitra, and Peter Wonka.
\newblock Styleflow: Attribute-conditioned exploration of stylegan-generated
  images using conditional continuous normalizing flows.
\newblock {\em ACM Trans. Graph.}, 40(3), May 2021.

\bibitem{geometric2018}
Hassan~Abu Alhaija, Siva~Karthik Mustikovela, Andreas Geiger, and Carsten
  Rother.
\newblock Geometric image synthesis.
\newblock {\em ACCV}, 2018.

\bibitem{BarronM15}
Jonathan~T. Barron and Jitendra Malik.
\newblock {Shape, Illumination, and Reflectance From Shading}.
\newblock {\em {IEEE} Transactions on Pattern Analysis and Machine
  Intelligence}, 2015.

\bibitem{barron2021mipnerf}
Jonathan~T. Barron, Ben Mildenhall, Matthew Tancik, Peter Hedman, Ricardo
  Martin-Brualla, and Pratul~P. Srinivasan.
\newblock Mip-nerf: A multiscale representation for anti-aliasing neural
  radiance fields, 2021.

\bibitem{binkowski2018demystifying}
Miko{\l}aj Bi{\'n}kowski, Danica~J Sutherland, Michael Arbel, and Arthur
  Gretton.
\newblock Demystifying mmd gans.
\newblock {\em arXiv preprint arXiv:1801.01401}, 2018.

\bibitem{Boss2020-NeRD}
Mark Boss, Raphael Braun, Varun Jampani, Jonathan~T. Barron, Ce Liu, and
  Hendrik~P.A. Lensch.
\newblock Nerd: Neural reflectance decomposition from image collections.
\newblock {\em ICCV}, 2020.

\bibitem{boss2021neuralpil}
Mark Boss, Varun Jampani, Raphael Braun, Ce Liu, Jonathan~T. Barron, and
  Hendrik~P.A. Lensch.
\newblock Neural-pil: Neural pre-integrated lighting for reflectance
  decomposition.
\newblock In {\em NeurIPS}, 2021.

\bibitem{biggan}
Andrew Brock, Jeff Donahue, and Karen Simonyan.
\newblock Large scale {GAN} training for high fidelity natural image synthesis.
\newblock In {\em ICLR}, 2019.

\bibitem{buehler2021varitex}
Marcel~C. Buehler, Abhimitra Meka, Gengyan Li, Thabo Beeler, and Otmar
  Hilliges.
\newblock Varitex: Variational neural face textures.
\newblock In {\em Proceedings of the IEEE/CVF International Conference on
  Computer Vision}, 2021.

\bibitem{Chan2021}
Eric~R. Chan, Connor~Z. Lin, Matthew~A. Chan, Koki Nagano, Boxiao Pan,
  Shalini~De Mello, Orazio Gallo, Leonidas Guibas, Jonathan Tremblay, Sameh
  Khamis, Tero Karras, and Gordon Wetzstein.
\newblock Efficient geometry-aware {3D} generative adversarial networks.
\newblock In {\em arXiv}, 2021.

\bibitem{chan2021pi}
Eric~R Chan, Marco Monteiro, Petr Kellnhofer, Jiajun Wu, and Gordon Wetzstein.
\newblock pi-gan: Periodic implicit generative adversarial networks for
  3d-aware image synthesis.
\newblock In {\em Proceedings of the IEEE/CVF Conference on Computer Vision and
  Pattern Recognition}, pages 5799--5809, 2021.

\bibitem{chen2021ngp}
Xuelin Chen, Daniel Cohen-Or, Baoquan Chen, and Niloy~J. Mitra.
\newblock Towards a neural graphics pipeline for controllable image generation.
\newblock {\em Computer Graphics Forum}, 40(2), 2021.

\bibitem{chogovadze2021controllable}
George Chogovadze, Rémi Pautrat, and Marc Pollefeys.
\newblock Controllable data augmentation through deep relighting, 2021.

\bibitem{choi2020stargan}
Yunjey Choi, Youngjung Uh, Jaejun Yoo, and Jung-Woo Ha.
\newblock Stargan v2: Diverse image synthesis for multiple domains, 2020.

\bibitem{deng2019arcface}
Jiankang Deng, Jia Guo, Niannan Xue, and Stefanos Zafeiriou.
\newblock Arcface: Additive angular margin loss for deep face recognition.
\newblock In {\em Proceedings of the IEEE/CVF Conference on Computer Vision and
  Pattern Recognition}, pages 4690--4699, 2019.

\bibitem{deng2021gram}
Yu Deng, Jiaolong Yang, Jianfeng Xiang, and Xin Tong.
\newblock Gram: Generative radiance manifolds for 3d-aware image generation.
\newblock {\em arXiv preprint arXiv:2112.08867}, 2021.

\bibitem{durugkar2017generative}
Ishan Durugkar, Ian Gemp, and Sridhar Mahadevan.
\newblock Generative multi-adversarial networks, 2017.

\bibitem{gadelha20173d}
Matheus Gadelha, Subhransu Maji, and Rui Wang.
\newblock 3d shape induction from 2d views of multiple objects.
\newblock In {\em 2017 International Conference on 3D Vision (3DV)}, pages
  402--411. IEEE, 2017.

\bibitem{GANs}
Ian Goodfellow, Jean Pouget-Abadie, Mehdi Mirza, Bing Xu, David Warde-Farley,
  Sherjil Ozair, Aaron Courville, and Yoshua Bengio.
\newblock Generative adversarial nets.
\newblock In {\em Advances in Neural Information Processing Systems}, 2014.

\bibitem{greene1986environment}
Ned Greene.
\newblock Environment mapping and other applications of world projections.
\newblock {\em IEEE Computer Graphics and Applications}, 6(11):21--29, 1986.

\bibitem{gu2021stylenerf}
Jiatao Gu, Lingjie Liu, Peng Wang, and Christian Theobalt.
\newblock Stylenerf: A style-based 3d-aware generator for high-resolution image
  synthesis.
\newblock {\em arXiv preprint arXiv:2110.08985}, 2021.

\bibitem{guo2019relightables}
Kaiwen Guo, Peter Lincoln, Philip Davidson, Jay Busch, Xueming Yu, Matt Whalen,
  Geoff Harvey, Sergio Orts-Escolano, Rohit Pandey, Jason Dourgarian, et~al.
\newblock The relightables: Volumetric performance capture of humans with
  realistic relighting.
\newblock {\em ACM Transactions on Graphics (TOG)}, 38(6):1--19, 2019.

\bibitem{heusel2017gans}
Martin Heusel, Hubert Ramsauer, Thomas Unterthiner, Bernhard Nessler, and Sepp
  Hochreiter.
\newblock Gans trained by a two time-scale update rule converge to a local nash
  equilibrium.
\newblock {\em Advances in neural information processing systems}, 30, 2017.

\bibitem{hong2021headnerf}
Yang Hong, Bo Peng, Haiyao Xiao, Ligang Liu, and Juyong Zhang.
\newblock Headnerf: A real-time nerf-based parametric head model.
\newblock {\em arXiv preprint arXiv:2112.05637}, 2021.

\bibitem{Kanamori:2018}
Yoshihiro Kanamori and Yuki Endo.
\newblock {Relighting Humans: Occlusion-Aware Inverse Rendering for Full-Body
  Human Images}.
\newblock {\em ACM Transactions Graphics (Proc. SIGGRAPH Asia)}, 2018.

\bibitem{karras2017progressive}
Tero Karras, Timo Aila, Samuli Laine, and Jaakko Lehtinen.
\newblock Progressive growing of gans for improved quality, stability, and
  variation.
\newblock {\em arXiv preprint arXiv:1710.10196}, 2017.

\bibitem{Karras2021}
Tero Karras, Miika Aittala, Samuli Laine, Erik H\"ark\"onen, Janne Hellsten,
  Jaakko Lehtinen, and Timo Aila.
\newblock Alias-free generative adversarial networks.
\newblock In {\em NeurIPS}, 2021.

\bibitem{karras2019style}
Tero Karras, Samuli Laine, and Timo Aila.
\newblock A style-based generator architecture for generative adversarial
  networks.
\newblock In {\em Proceedings of the IEEE/CVF Conference on Computer Vision and
  Pattern Recognition}, pages 4401--4410, 2019.

\bibitem{Karras2019stylegan2}
Tero Karras, Samuli Laine, Miika Aittala, Janne Hellsten, Jaakko Lehtinen, and
  Timo Aila.
\newblock Analyzing and improving the image quality of {StyleGAN}.
\newblock In {\em Proc. CVPR}, 2020.

\bibitem{Liao2020CVPR}
Yiyi Liao, Katja Schwarz, Lars Mescheder, and Andreas Geiger.
\newblock Towards unsupervised learning of generative models for 3d
  controllable image synthesis.
\newblock In {\em Proceedings IEEE Conf. on Computer Vision and Pattern
  Recognition (CVPR)}, 2020.

\bibitem{liu2015faceattributes}
Ziwei Liu, Ping Luo, Xiaogang Wang, and Xiaoou Tang.
\newblock Deep learning face attributes in the wild.
\newblock In {\em Proceedings of International Conference on Computer Vision
  (ICCV)}, December 2015.

\bibitem{mallikarjun2021photoapp}
B~R Mallikarjun, Ayush Tewari, Abdallah Dib, Tim Weyrich, Bernd Bickel,
  Hans-Peter Seidel, Hanspeter Pfister, Wojciech Matusik, Louis Chevallier,
  Mohamed Elgharib, et~al.
\newblock Photoapp: Photorealistic appearance editing of head portraits.
\newblock {\em ACM Transactions on Graphics}, 40(4):1--16, 2021.

\bibitem{meka:2017}
Abhimitra Meka, Gereon Fox, Michael Zollh{\"o}fer, Christian Richardt, and
  Christian Theobalt.
\newblock {Live User-Guided Intrinsic Video for Static Scene}.
\newblock {\em IEEE Transactions on Visualization and Computer Graphics}, 2017.

\bibitem{Meka:2018}
Abhimitra Meka, Maxim Maximov, Michael Zollhoefer, Avishek Chatterjee,
  Hans-Peter Seidel, Christian Richardt, and Christian Theobalt.
\newblock {LIME: Live Intrinsic Material Estimation}.
\newblock In {\em Proc. Computer Vision and Pattern Recognition}, 2018.

\bibitem{deep_relightables}
Abhimitra Meka, Rohit Pandey, Christian H\"{a}ne, Sergio Orts-Escolano, Peter
  Barnum, Philip David-Son, Daniel Erickson, Yinda Zhang, Jonathan Taylor,
  Sofien Bouaziz, Chloe LeGendre, Wan-Chun Ma, Ryan Overbeck, Thabo Beeler,
  Paul Debevec, Shahram Izadi, Christian Theobalt, Christoph Rhemann, and Sean
  Fanello.
\newblock Deep relightable textures: Volumetric performance capture with neural
  rendering.
\newblock {\em ACM Transactions on Graphics}, 2020.

\bibitem{mildenhall2020nerf}
Ben Mildenhall, Pratul~P Srinivasan, Matthew Tancik, Jonathan~T Barron, Ravi
  Ramamoorthi, and Ren Ng.
\newblock Nerf: Representing scenes as neural radiance fields for view
  synthesis.
\newblock In {\em European conference on computer vision}, pages 405--421.
  Springer, 2020.

\bibitem{miller1984illumination}
Gene~S Miller and CR Hoffman.
\newblock Illumination and reflection maps.
\newblock In {\em ACM SIGGRAPH}, 1984.

\bibitem{mordido2020dropoutgan}
Gonçalo Mordido, Haojin Yang, and Christoph Meinel.
\newblock Dropout-gan: Learning from a dynamic ensemble of discriminators,
  2018.

\bibitem{nestmeyer2020learning}
Thomas Nestmeyer, Jean-François Lalonde, Iain Matthews, and Andreas~M.
  Lehrmann.
\newblock Learning physics-guided face relighting under directional light.
\newblock In {\em CVPR}, 2020.

\bibitem{hologan19}
Thu Nguyen-Phuoc, Chuan Li, Lucas Theis, Christian Richardt, and Yong-Liang
  Yang.
\newblock Hologan: Unsupervised learning of 3d representations from natural
  images.
\newblock In {\em The IEEE International Conference on Computer Vision (ICCV)},
  Nov 2019.

\bibitem{niemeyer2021giraffe}
Michael Niemeyer and Andreas Geiger.
\newblock Giraffe: Representing scenes as compositional generative neural
  feature fields.
\newblock In {\em Proceedings of the IEEE/CVF Conference on Computer Vision and
  Pattern Recognition}, pages 11453--11464, 2021.

\bibitem{or2021stylesdf}
Roy Or-El, Xuan Luo, Mengyi Shan, Eli Shechtman, Jeong~Joon Park, and Ira
  Kemelmacher-Shlizerman.
\newblock Stylesdf: High-resolution 3d-consistent image and geometry
  generation.
\newblock {\em arXiv e-prints}, pages arXiv--2112, 2021.

\bibitem{pan2021shadegan}
Xingang Pan, Xudong Xu, Chen~Change Loy, Christian Theobalt, and Bo Dai.
\newblock A shading-guided generative implicit model for shape-accurate
  3d-aware image synthesis.
\newblock In {\em NeurIPS}, 2021.

\bibitem{pandey2021total}
Rohit Pandey, Sergio~Orts Escolano, Chloe Legendre, Christian Haene, Sofien
  Bouaziz, Christoph Rhemann, Paul Debevec, and Sean Fanello.
\newblock Total relighting: learning to relight portraits for background
  replacement.
\newblock {\em ACM Transactions on Graphics (TOG)}, 40(4):1--21, 2021.

\bibitem{parmar2021cleanfid}
Gaurav Parmar, Richard Zhang, and Jun-Yan Zhu.
\newblock On buggy resizing libraries and surprising subtleties in fid
  calculation.
\newblock {\em arXiv preprint arXiv:2104.11222}, 2021.

\bibitem{phong1975illumination}
Bui~Tuong Phong.
\newblock Illumination for computer generated pictures.
\newblock {\em Communications of the ACM}, 18(6):311--317, 1975.

\bibitem{phuoc2020blockgan}
Thu~Nguyen Phuoc, Christian Richardt, Long Mai, Yongliang Yang, and Niloy~J
  Mitra.
\newblock Blockgan: Learning 3d object-aware scene representations from
  unlabelled images.
\newblock In {\em NeurIPS 2020: Conference on Neural Information Processing
  Systems}, 2020.

\bibitem{ramamoorthi2001efficient}
Ravi Ramamoorthi and Pat Hanrahan.
\newblock An efficient representation for irradiance environment maps.
\newblock In {\em Proceedings of the 28th annual conference on Computer
  graphics and interactive techniques}, pages 497--500, 2001.

\bibitem{Ren:2015}
Peiran Ren, Yue Dong, Stephen Lin, Xin Tong, and Baining Guo.
\newblock {Image Based Relighting Using Neural Networks}.
\newblock {\em ACM Transactions on Graphics}, 2015.

\bibitem{ronneberger2015u}
Olaf Ronneberger, Philipp Fischer, and Thomas Brox.
\newblock U-net: Convolutional networks for biomedical image segmentation.
\newblock In {\em International Conference on Medical image computing and
  computer-assisted intervention}, pages 234--241. Springer, 2015.

\bibitem{roessler2019faceforensics}
Andreas R\"ossler, Davide Cozzolino, Luisa Verdoliva, Christian Riess, Justus
  Thies, and Matthias Nie{\ss}ner.
\newblock Faceforensics++: Learning to detect manipulated facial images.
\newblock In {\em ICCV 2019}, 2019.

\bibitem{saito2020pifuhd}
Shunsuke Saito, Tomas Simon, Jason Saragih, and Hanbyul Joo.
\newblock Pifuhd: Multi-level pixel-aligned implicit function for
  high-resolution 3d human digitization.
\newblock In {\em Proceedings of the IEEE Conference on Computer Vision and
  Pattern Recognition}, June 2020.

\bibitem{schwarz2021graf}
Katja Schwarz, Yiyi Liao, Michael Niemeyer, and Andreas Geiger.
\newblock Graf: Generative radiance fields for 3d-aware image synthesis, 2020.

\bibitem{shi2016real}
Wenzhe Shi, Jose Caballero, Ferenc Husz{\'a}r, Johannes Totz, Andrew~P Aitken,
  Rob Bishop, Daniel Rueckert, and Zehan Wang.
\newblock Real-time single image and video super-resolution using an efficient
  sub-pixel convolutional neural network.
\newblock In {\em Proceedings of the IEEE conference on computer vision and
  pattern recognition}, pages 1874--1883, 2016.

\bibitem{szabo2019unsupervised}
Attila Szab{\'o}, Givi Meishvili, and Paolo Favaro.
\newblock Unsupervised generative 3d shape learning from natural images.
\newblock {\em arXiv preprint arXiv:1910.00287}, 2019.

\bibitem{tajima2021relighting}
Daichi Tajima, Yoshihiro Kanamori, and Yuki Endo.
\newblock Relighting humans in the wild: Monocular full-body human relighting
  with domain adaptation, 2021.

\bibitem{tan2021humangps}
Feitong Tan, Danhang Tang, Dou Mingsong, Guo Kaiwen, Rohit Pandey, Cem Keskin,
  Ruofei Du, Deqing Sun, Sofien Bouaziz, Sean Fanello, Ping Tan, and Yinda
  Zhang.
\newblock Humangps: Geodesic preserving feature for dense human
  correspondences.
\newblock In {\em Proceedings of the IEEE/CVF Conference on Computer Vision and
  Pattern Recognition (CVPR)}, June 2021.

\bibitem{tancik2021}
Matthew Tancik, Pratul~P. Srinivasan, Ben Mildenhall, Sara Fridovich{-}Keil,
  Nithin Raghavan, Utkarsh Singhal, Ravi Ramamoorthi, Jonathan~T. Barron, and
  Ren Ng.
\newblock Fourier features let networks learn high frequency functions in low
  dimensional domains.
\newblock 2021.

\bibitem{tewari2020stylerig}
Ayush Tewari, Mohamed Elgharib, Gaurav Bharaj, Florian Bernard, Hans-Peter
  Seidel, Patrick P{\'e}rez, Michael Z{\"o}llhofer, and Christian Theobalt.
\newblock Stylerig: Rigging stylegan for 3d control over portrait images, cvpr
  2020.
\newblock In {\em {IEEE} Conference on Computer Vision and Pattern Recognition
  (CVPR)}. {IEEE}, june 2020.

\bibitem{tewari2020state}
Ayush Tewari, Ohad Fried, Justus Thies, Vincent Sitzmann, Stephen Lombardi,
  Kalyan Sunkavalli, Ricardo Martin-Brualla, Tomas Simon, Jason Saragih,
  Matthias Nießner, Rohit Pandey, Sean Fanello, Gordon Wetzstein, Jun-Yan Zhu,
  Christian Theobalt, Maneesh Agrawala, Eli Shechtman, Dan~B Goldman, and
  Michael Zollhoefer.
\newblock State of the art on neural rendering.
\newblock In {\em Eurographics}, 2020.

\bibitem{deferred}
Justus Thies, Michael Zollh\"{o}fer, and Matthias Niessner.
\newblock Deferred neural rendering: Image synthesis using neural textures.
\newblock {\em SIGGRAPH and ACM TOG}, 2019.

\bibitem{tolosana20}
Ruben Tolosana, Ruben Vera-Rodriguez, Julian Fierrez, Aythami Morales, and
  Javier Ortega-Garcia.
\newblock Deepfakes and beyond: A survey of face manipulation and fake
  detection.
\newblock {\em Information Fusion}, 2020.

\bibitem{Wang20}
Zhibo Wang, Xin Yu, Ming Lu, Quan Wang, Chen Qian, and Feng Xu.
\newblock Single image portrait relighting via explicit multiple reflectance
  channel modeling.
\newblock {\em ACM SIGGRAPH Asia and Transactions on Graphics}, 2020.

\bibitem{wood2021fake}
Erroll Wood, Tadas Baltru\v{s}aitis, Charlie Hewitt, Sebastian Dziadzio,
  Matthew Johnson, Virginia Estellers, Thomas~J. Cashman, and Jamie Shotton.
\newblock Fake it till you make it: Face analysis in the wild using synthetic
  data alone, 2021.

\bibitem{xu2018deep}
Zexiang Xu, Kalyan Sunkavalli, Sunil Hadap, and Ravi Ramamoorthi.
\newblock Deep image-based relighting from optimal sparse samples.
\newblock {\em ACM Transactions on Graphics}, 2018.

\bibitem{HDRIHaven}
Greg Zaal, Sergej Majboroda, and Andreas Mischok.
\newblock Hdri haven.
\newblock \url{https://www.hdrihaven.com/}, 2020.
\newblock Accessed: 2021-11-13.

\bibitem{pmlr-v97-zhang19d}
Han Zhang, Ian Goodfellow, Dimitris Metaxas, and Augustus Odena.
\newblock Self-attention generative adversarial networks.
\newblock In {\em International Conference on Machine Learning}, 2019.

\bibitem{zhang2021ners}
Jason~Y. Zhang, Gengshan Yang, Shubham Tulsiani, and Deva Ramanan.
\newblock {NeRS}: Neural reflectance surfaces for sparse-view 3d reconstruction
  in the wild.
\newblock In {\em Conference on Neural Information Processing Systems}, 2021.

\bibitem{zhang2019making}
Richard Zhang.
\newblock Making convolutional networks shift-invariant again.
\newblock In {\em International conference on machine learning}, pages
  7324--7334. PMLR, 2019.

\bibitem{zhang2021nerfactor}
Xiuming Zhang, Pratul~P Srinivasan, Boyang Deng, Paul Debevec, William~T
  Freeman, and Jonathan~T Barron.
\newblock Nerfactor: Neural factorization of shape and reflectance under an
  unknown illumination.
\newblock {\em arXiv preprint arXiv:2106.01970}, 2021.

\bibitem{zhou2021cips}
Peng Zhou, Lingxi Xie, Bingbing Ni, and Qi Tian.
\newblock Cips-3d: A 3d-aware generator of gans based on
  conditionally-independent pixel synthesis.
\newblock {\em arXiv preprint arXiv:2110.09788}, 2021.

\bibitem{visual2018}
Jun-Yan Zhu, Zhoutong Zhang, Chengkai Zhang, Jiajun Wu, Antonio Torralba, Josh
  Tenenbaum, and Bill Freeman.
\newblock Visual object networks: Image generation with disentangled 3d
  representations.
\newblock In {\em Advances in Neural Information Processing Systems}, 2018.

\bibitem{zhu2020simpose}
Tyler Zhu, Per Karlsson, and Christoph Bregler.
\newblock Simpose: Effectively learning densepose and surface normals of people
  from simulated data.
\newblock In {\em European Conference on Computer Vision}, pages 225--242.
  Springer, 2020.

\end{thebibliography}
}
\clearpage
\appendix
\section*{Appendix}

\begin{figure*}
    \centering
    \includegraphics[width=\linewidth]{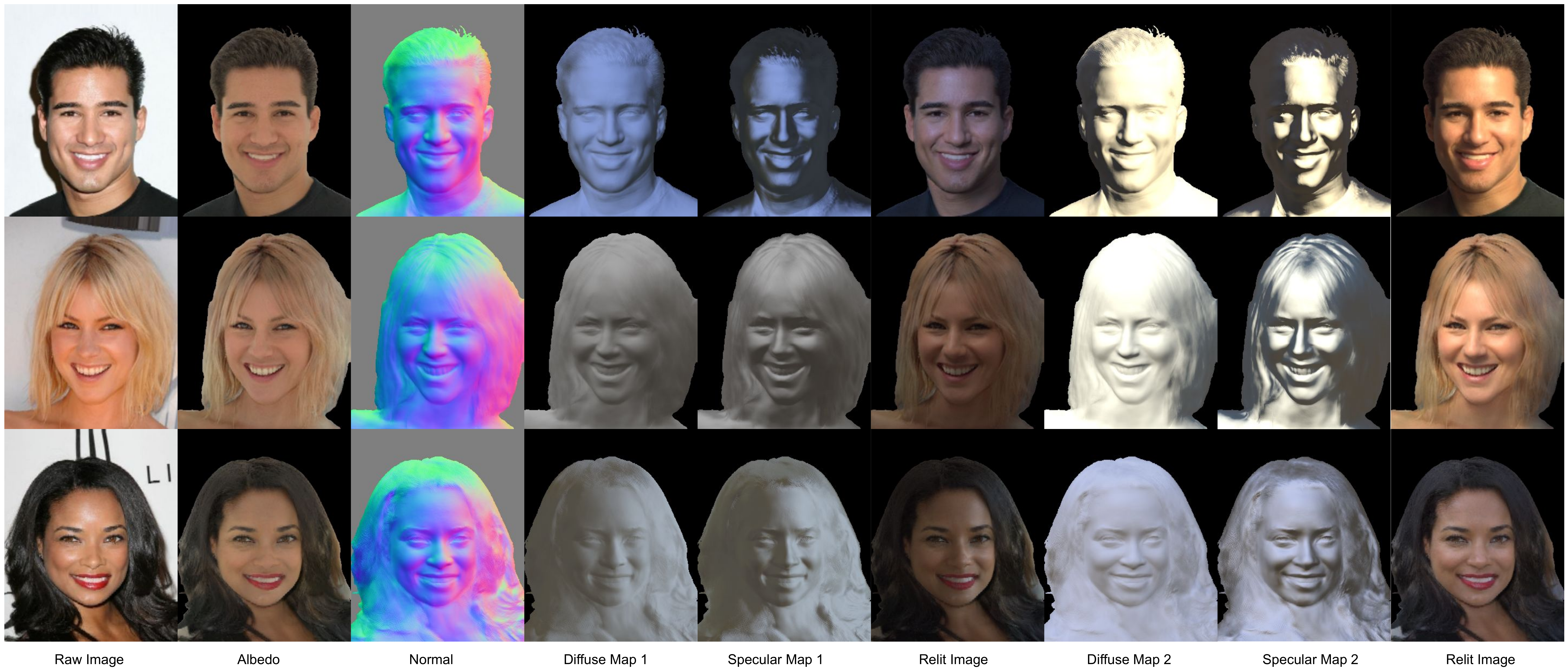}
    \caption{Relighting augmentation on CelebA \cite{liu2015faceattributes} using \cite{pandey2021total} to generate albedo, normal, shading, and relit images with different HDRI Relighting, which supervise the training via adversarial losses.}
    \label{fig:celeba_dataset}
\end{figure*}

\begin{figure*}
    \centering
    \includegraphics[width=\linewidth]{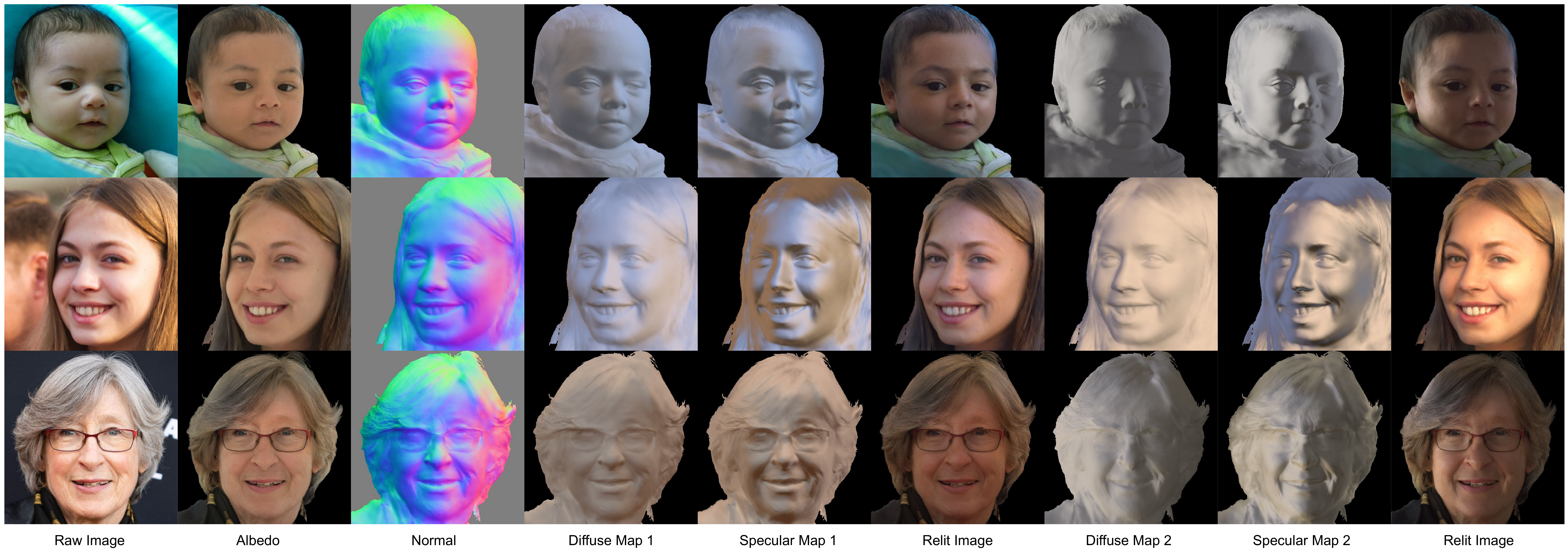}
    \caption{Relighting augmentation on FFHQ \cite{karras2019style} using \cite{pandey2021total}  to generate albedo, normal, shading, and relit images with different HDRI Relighting, which supervise the training via adversarial losses.}
    \label{fig:ffhq_dataset}
\end{figure*}

In this appendix, we provide more details regarding the proposed data augmentation strategy, network architecture and additional results. Finally, we also discuss the limitations of the model.
We also provide a supplementary HTML page showing animated results of generated face under various camera viewpoints and environmental illuminations.

\section{Data Augmentation via Portrait Relighting}
We provide additional information regarding our data augmentation strategy which uses the portrait relighting method of \cite{pandey2021total} to produce pseudo ground truth albedo, normals, a relit image and the associated light maps (diffuse and specular components) on the CelebA \cite{liu2015faceattributes} and the FFHQ \cite{karras2019style} datasets. Specifically, we generate $5$ and $10$ relit images for each image in CelebA and FFHQ datasets. The HDRI map is randomly sampled from a collection of $400$ maps sourced from public repository \cite{HDRIHaven} and randomly rotated horizontally. We show more example images of the augmented CelebA images and FFHQ images in Figure \ref{fig:celeba_dataset} and Figure \ref{fig:ffhq_dataset}. For each identity, we visualize the relit image and the associated light maps with two different HDRI images.

\begin{figure*}
    \centering
    \includegraphics[width=\linewidth]{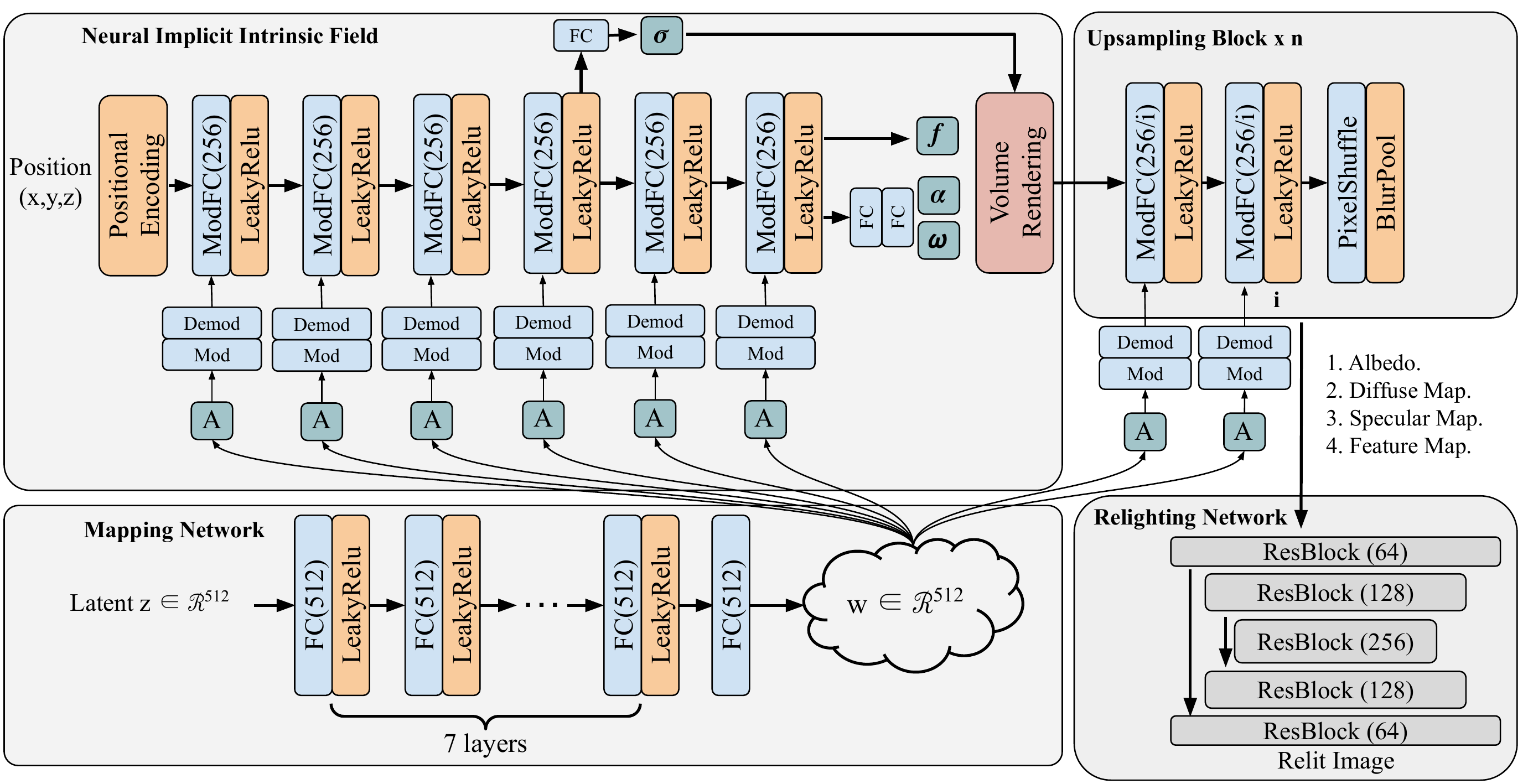}
    \caption{Proposed architecture of our neural generator, which consists of a neural implicit intrinsic field network, upsampling blocks, a relighting network and a mapping network.}
    \label{fig:architecture}
\end{figure*}

\section{Network Architecture}
The details of the proposed architecture are shown in Figure \ref{fig:architecture}.  As detailed in the main paper, the framework consists of four modules: a neural implicit intrinsic field (NeIIF) network, upsampling blocks, a relighting network and a mapping network. Similar to StyleGAN2 \cite{Karras2019stylegan2}, the mapping network consists of 8 fully-connected layers with 512 units, that maps the latent code to a style vector. The output vectors are then broadcast to every fully-connected layer in the NeIIF network and the upsampling blocks. For each vector, there is an affine transformation layer to map it to an affine-transformed style, which is used to modulate the feature maps of the NeIIF network and upsampling block. The NeIIF network consist of a positional encoder (the Fourier feature dimension is set to 10) and a 6-layer MLP with 256 units. The feature maps of each fully-connected layer are modulated by an affine transformation from the mapping network. Each upsampling block consists of two fully-connected layers modulated by the latent code z, a pixelshuffle upsampler and a BlurPool with stride 1 which increases the resolution by 2x. The relighting network is a residual U-Net with skip connections.

\section{Additional Results}
Here we show more results from our method.

\subsection{Intermediate Intrinsic Images}
\begin{figure*}
    \centering
    \includegraphics[width=\linewidth]{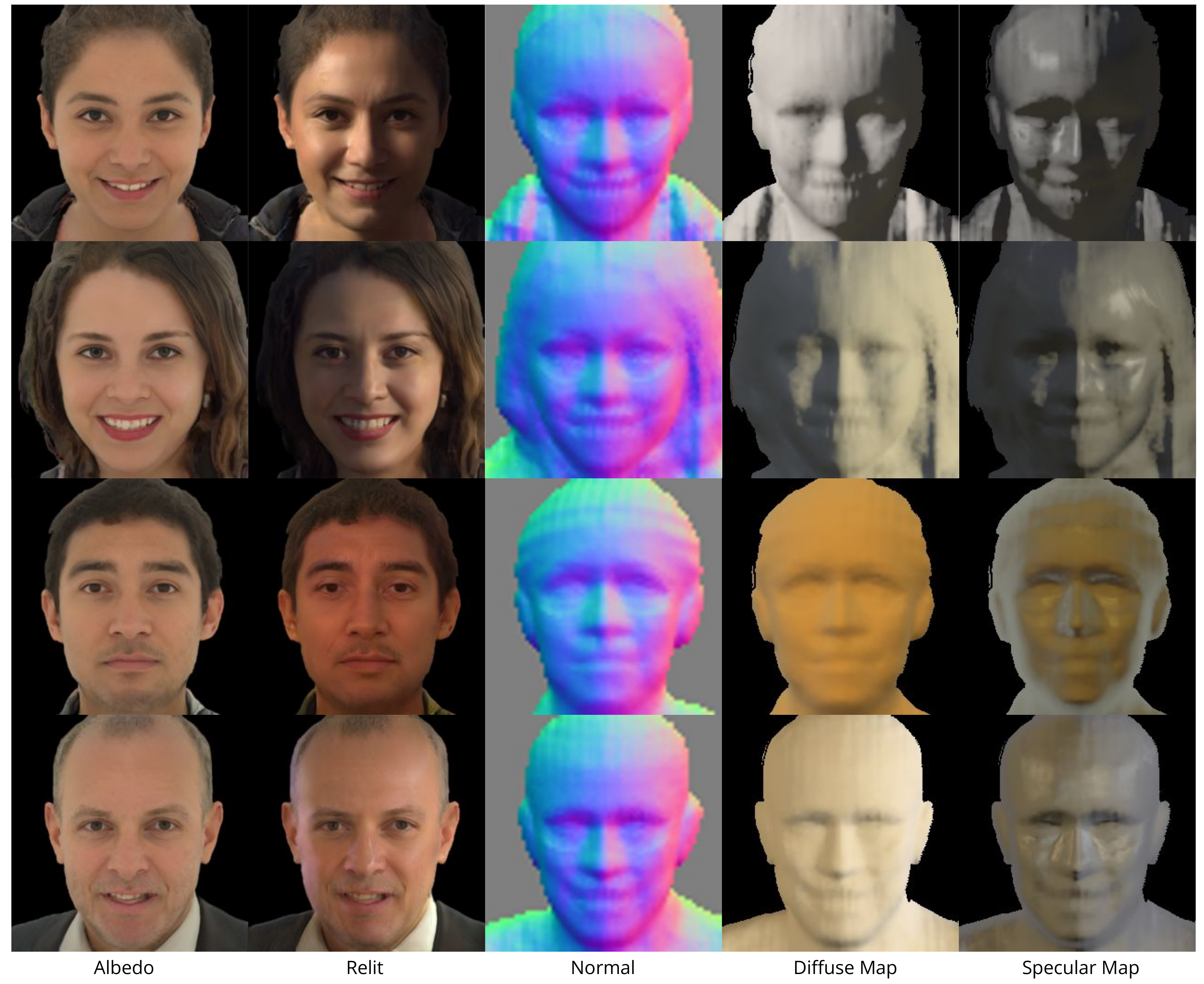}
    \caption{Results of intermediate intrinsic images from our model trained on FFHQ \cite{karras2019style}. From left to right, we show the albedo, relit image, normal map, diffuse map and specular map.}
    \label{fig:image_decomposition}
\end{figure*}

In Figure \ref{fig:image_decomposition}, we visualize the albedo, relit image, normal map, diffuse map and specular map from our generator trained on FFHQ dataset.
Note that since normal and shading maps are directly generated from the neural implicit field, we render them in low resolution for efficient training, which is a strategy adopted and demonstrated to be successful by other works \cite{gu2021stylenerf}.
We then rely on the generated feature map $F$ to produce high frequency details in the albedo and the final relit results.

\subsection{Relighting Accuracy}
\begin{figure*}
    \centering
    \includegraphics[width=\linewidth]{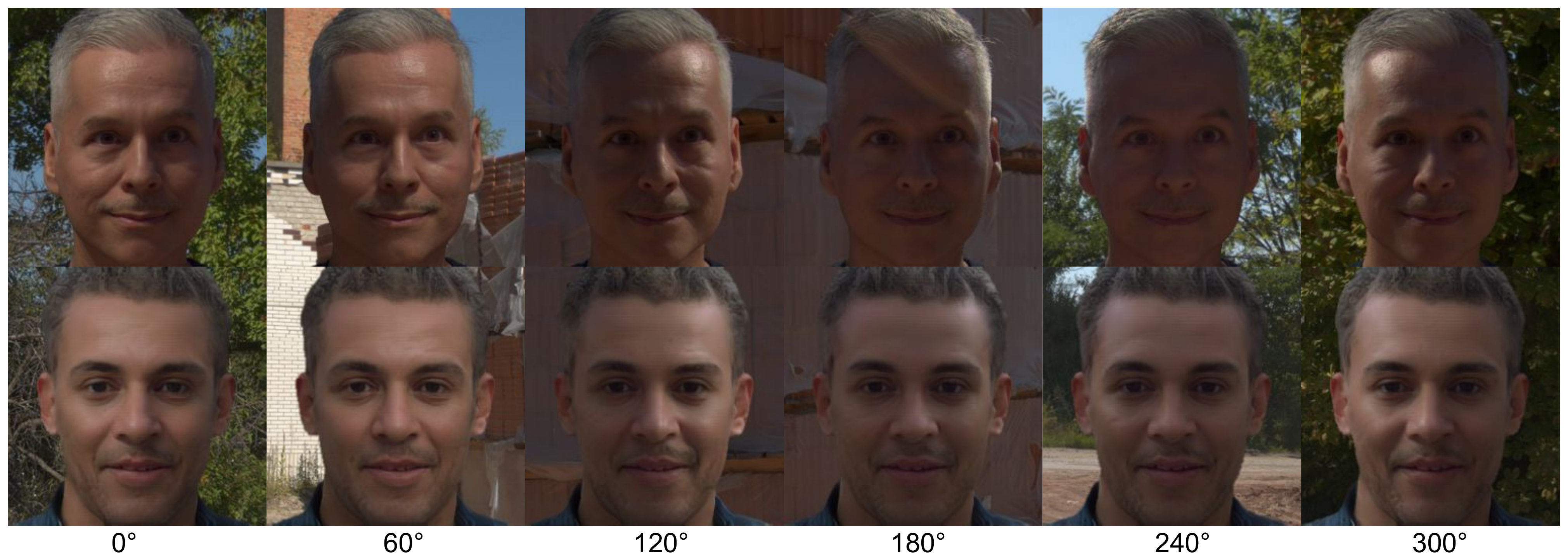}
    \caption{We compare our relighting result to image based relighting (IBR) using a Light Stage \cite{guo2019relightables} with the same HDRI illumination. Note that our method produces consistent and plausible shading, soft shadows and specularities.}
    \label{fig:comp_to_lightstage}
\end{figure*}

We show a qualitative comparison of our relighting method with environmental relighting of a real person captured in a dense high-resolution Light Stage in Figure \ref{fig:comp_to_lightstage}, which is very close to ground truth relighting. Note that as the environment map rotates, our method produces plausible shadows and specularities that spatially match the pseudo-ground-truth setup, indicating that our underlying 3D volumetric geometry and skin reflectance is stable. While there is some dampening of specularities and cast shadows, the overall identity of the generated person is well preserved, which is a significant improvement over the state-of-the-art \cite{pan2021shadegan}.

\begin{figure*}
    \centering
    \includegraphics[width=\linewidth]{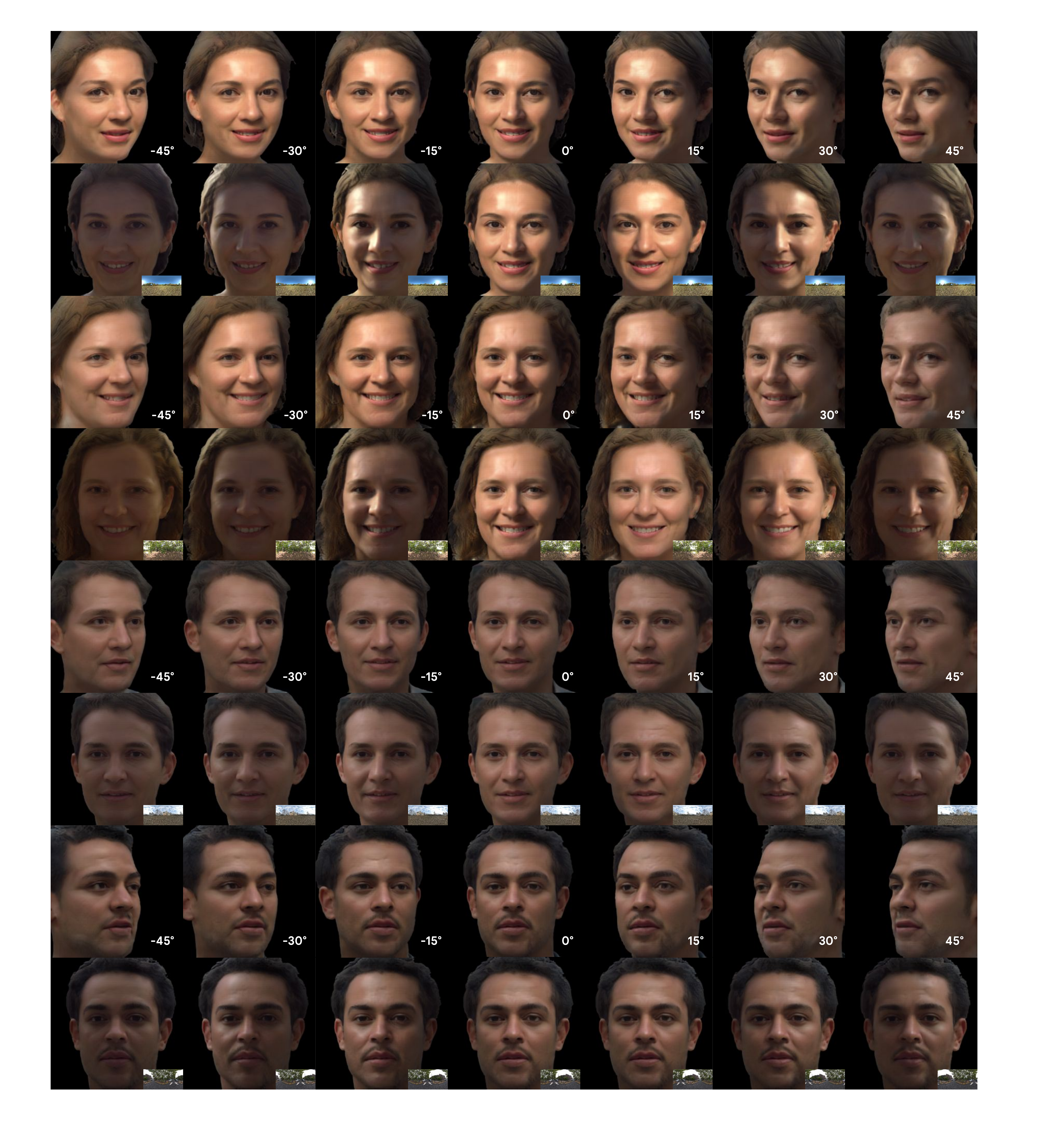}
    \caption{More synthesized images under rotating camera or rotating lighting. Note the relighting consistency and view-dependent effects.}
    \label{fig:rot_light_camera}
\end{figure*}




\subsection{Rotate Camera and Lighting}
We show more subjects generated from the model trained on the FFHQ dataset with randomly sampled latent codes in Figure  \ref{fig:rot_light_camera}. 
For each identity (\ie latent code), we show the rendering under the same HDRI map but different camera pose, and the rendering under a fixed camera pose with rotating HDRI map.
The results indicate that our method provides controllability over camera viewpoint and illumination, and deliver faithful rendering results.

\section{Animated Results in Companion HTML  Page}
We provide a supplementary HTML page to show animated rendering results.
Please open with your local browser.
In the HTML page, we show 1) our intermediate intrinsic results and final relighting results in a continuous camera trajectory, 2) comparison on the relighting faithfulness to ShadeGAN \cite{pan2021shadegan} under rotating HDRI, using image based relighting with a Light Stage \cite{guo2019relightables} as the reference, 3) relighting of the same or different subjects under same or different environment map, 4) multi-view synthesis, 5) a comparison on albedo stability with the baseline of piGAN \cite{chan2021pi} + TR \cite{pandey2021total}.

\section{Limitations}
Although the proposed approach is a step forward towards generative relightable 3D faces, it still has limitations. First, it lacks high frequency details on geometry and albedo when rendered at high resolutions (see Figure \ref{fig:image_decomposition}), despite our high quality supervision: we believe that using intuition from previous work \cite{karras2019style,Karras2019stylegan2,biggan} could help address this. 

At more extreme viewpoint changes, the identity similarity scores drop as demonstrated in Table 1 in the main paper, indicating that stronger pose/viewpoint changes may result in distortion of identity. This is likely due to skewed distribution of our in-the-wild training data which is mostly frontal, with very few side facing views. We believe that this can be improved by more carefully curating the training data using importance sampling to have a more even distribution of facial poses. Yet, please note that our method outperforms other state-of-the-art 3D synthesis methods \cite{pan2021shadegan,chan2021pi}, which in turn are significantly better than 2D based generative view synthesis methods \cite{tewari2020stylerig, mallikarjun2021photoapp, 10.1145/3447648}.

Furthermore, aliasing effects are noticeable when changing viewpoints especially around the teeth and hair. An approach similar to \cite{barron2021mipnerf} could potentially mitigate these effects.

Additionally, although our model shows impressive relighting results, it still cannot capture the same details of specular highlights when compared to image based relighting using a Light Stage as shown in Figure \ref{fig:comp_to_lightstage}. Additional losses that focus on specularities may help mitigate this issue.

Finally, the lack of supervision on the actual facial expression, makes the model unconstrained, leading to different face gestures when changing the viewpoint (see animated results in the provided HTML page). Adding semantic information such as keypoints or per-pixel labels could be an effective way to enable control over the expressions and ensure more consistency across views.

\end{document}